\documentclass{article}

\usepackage[preprint]{neurips_2026}

\usepackage[utf8]{inputenc} 
\usepackage[T1]{fontenc}    
\usepackage{hyperref}       
\usepackage{url}            
\usepackage{booktabs}       
\usepackage{amsfonts}       
\usepackage{amssymb}        
\usepackage{nicefrac}       
\usepackage{microtype}      
\usepackage{xcolor}         
\usepackage{amsmath}
\usepackage{graphicx}       
\usepackage{makecell}       
\usepackage{colortbl}       
\usepackage{wrapfig}        
\newcommand{\rankone}[1]{\cellcolor{red!30}#1}
\newcommand{\ranktwo}[1]{\cellcolor{orange!40}#1}
\newcommand{\rankthree}[1]{\cellcolor{yellow!50}#1}

\title{Analogical Trajectory Transfer}

\author{%
Junho Kim\textsuperscript{1}, Eun Sun Lee\textsuperscript{1}, Gwangtak Bae\textsuperscript{1},  Seunggu Kang\textsuperscript{1}, and Young Min Kim\textsuperscript{1, 2}
\and {\small \phantom{ }} \vspace{-1em}\\
\textsuperscript{1} {Dept. of Electrical and Computer Engineering, Seoul National University} \\
\textsuperscript{2} {Interdisciplinary Program in Artificial Intelligence and INMC, Seoul National University} \\
{\tt\small \{82magnolia, tak3452, eunsunlee, youngmin.kim\}@snu.ac.kr}}

\begin{document}

\maketitle

\begin{abstract}
We study \textit{analogical trajectory transfer}, where the goal is to translate motion trajectories in one 3D environment to a semantically analogous location in another.
Such a capacity would enable machines to perform analogical spatial reasoning, with applications in AR/VR co-presence, content creation, and robotics.
However, even semantically similar scenes can still differ substantially in object placement, scale, and layout, so naively matching semantics leads to collisions or geometric distortions.
Furthermore, finding where each trajectory point should transfer to has a large search space, as the mapping must preserve semantics and functionality without tearing the trajectory apart or causing collisions.
Our key insight is to decompose the problem into spatially segregated subproblems and merge their solutions to produce semantically consistent and spatially coherent transfers.
Specifically, we partition scenes into object-centric clusters and estimate cross-scene mappings via hierarchical smooth map prediction, using 3D foundation model features that encode contextual information from object and open-space arrangements.
We then combinatorially assemble the per-cluster maps into an initial transfer and refine the result to remove collisions and distortions, yielding a spatially coherent trajectory.
Our method does not require training, attains a fast runtime around 0.6 seconds, and outperforms baselines based on LLMs, VLMs, and scene graph matching.
We further showcase applications in virtual co-presence, multi-trajectory transfer, camera transfer, and human-to-robot motion transfer, which indicates the broad applicability of our work to AR/VR and robotics.
\end{abstract}

\section{Introduction}
Many real-world activities involve rich, structured interactions with 3D environments, where both \emph{what} actions are performed and \emph{where} they occur are tightly coupled to the surrounding scene.
For example, preparing a meal requires moving between storage, preparation, and cooking regions in a specific sequence dictated by the kitchen's layout, while household routines such as doing laundry involve collecting clothes, transferring them to the washing area, and finally storing them in the closet.
To enable machines to perform such order-critical tasks in novel, unseen environments, a key capacity is the ability to translate trajectories observed in one environment into semantically \emph{analogous} trajectories in another; this transfer should preserve functional intent, respect geometric constraints, and maintain a coherent ordering of interactions.
In this paper, we term this capacity \emph{analogical trajectory transfer}, and argue that it serves as a crucial component for generalizing structured, scene-dependent activities across diverse 3D environments.

\begin{figure*}[t]
  \centering
    \includegraphics[width=\linewidth]{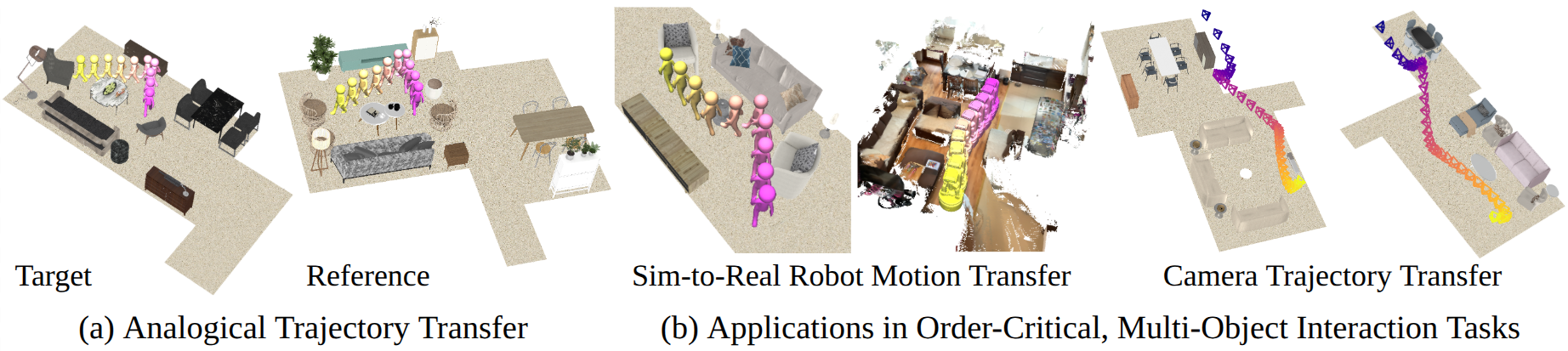}
   \caption{The goal of \emph{analogical trajectory transfer} is to map trajectory points from one scene to semantically and spatially \emph{analogous} locations in another, enabling applications in order-critical, multi-object interaction tasks such as sim-to-real robot motion transfer and camera trajectory transfer.}
   \label{fig:teaser}
   \vspace{-1em}
\end{figure*}

Achieving analogical trajectory transfer is far from straightforward, with key challenges lying on the modeling, data, and efficiency requirements of the task.
Since even scenes with similar functions and semantics can vary significantly in object appearance, layout, and arrangement, identifying corresponding locations for each trajectory point in a new scene entails a vast search space, further complicated by physical constraints such as collision avoidance.
Addressing this challenge requires understanding rich spatial context, as trajectories are shaped not only by individual objects but by their arrangement and the structure of surrounding free space.
Consequently, naively relying on object semantics alone often leads to collisions and trajectory distortions, as illustrated in Figure~\ref{fig:qualitative}.
The difficulty compounds when multiple agents are involved, as their trajectories must be transferred simultaneously while preserving pairwise interactions such as relative distances, which is difficult to attain with existing approaches.
While one could instead train a neural network that directly predicts analogical trajectory transfer (i.e., \emph{trajectory-in, trajectory-out}), collecting ground-truth trajectory pairs for supervision is non-trivial.
At the same time, many robotics and AR/VR applications require fast trajectory generation to support responsive interaction in dynamic environments.
As a result, producing transfers that are both semantically consistent and spatially coherent, while remaining computationally efficient and training-free, poses a significant challenge.

To address these challenges, we propose a hierarchical map prediction framework that leverages 3D foundation models~\citep{sonata} for analogical trajectory transfer.
The foundation model features capture rich scene context in 3D that goes beyond pairwise object relations by holistically encoding object semantics and spatial arrangements.
Further, the features are robust to sensor noise and appearance variations which enables reliable transfer in noisy real-world scans and sim-to-real scenarios involving large domain gaps.
Critically, we observe that these features are usable for our task without additional training, which enables our method to operate directly on raw scene pairs without any supervision.
Based on these features, our key idea is to adopt a \emph{divide-and-conquer strategy} to cope with the large search space: we partition scenes into object-centric clusters, find cluster-level matches and smooth maps, and assemble the maps followed by refinement to produce the final transfer.
To ensure semantic consistency, we align 3D foundation model features during cluster-level matching and initial smooth map estimation.
Specifically, we build object-level graphs that express higher-order relational context and holistically match them by comparing 3D foundation model features, followed by per-cluster smooth map fitting based on dense point-level feature comparison.
To ensure spatial coherence, we combinatorially search for smooth map combinations that minimize collisions and distortions while maintaining consistent 3D foundation model features, and further refine the transferred trajectory to remove any remaining collision and distortion artifacts.

We evaluate our method on synthetic scenes from 3D-FRONT, real-world noisy scans from ARKitScenes, and Sim2Real pairs across both datasets.
Our method outperforms baselines based on LLMs, VLMs, scene graphs, and prior methods for analogical spatial reasoning in both synthetic and real-world settings .
We further showcase applications in multi-trajectory transfer for AR/VR content creation, sim-to-real transfer of human motion to robot actions, virtual co-presence, and camera trajectory transfer.

To summarize, our key contributions are as follows: (i) we propose the \textit{analogical trajectory transfer} task and a solution that jointly considers semantic consistency and spatial coherence through a hierarchical cluster-level map finding and refinement framework; (ii) we design our framework using 3D foundation models without additional training and maintain fast inference suitable for interactive applications; (iii) we demonstrate applications for AR/VR and robotics that involve long, scene-scale trajectories and multi-agent interactions in complex indoor scenes.

\vspace{-1em}
\section{Related Work}

\textbf{3D Scene Understanding.}
Understanding objects, open spaces, and their relationships has been a long-standing problem in 3D computer vision.
Prior works have mainly focused on discovering entity-wise~\citep{point_contrast,pointnet,mask3d,openmask3d,openscene} or pairwise concepts~\citep{open3dsg,relation_field,td_scene_graph_armeni,clio,conceptgraphs,function_scene_graph} from 3D scenes.
At the entity level, methods recover object semantic and instance labels by aggregating 2D foundation model features to 3D~\citep{dino_in_the_room,sam_3d_seg,clipfields} or using multi-modal LLMs~\citep{openscene,openmask3d}.
At the pairwise level, works predict object-object relationships~\citep{open3dsg,td_scene_graph_armeni,sgrec3d,scene_graph_fusion} or human-object affordances~\citep{affordance_net,mo2021o2oafford,3daffordancellm,affordance_diffusion,affordance_learning_coma} from scene geometry and semantics.
Our task instead focuses on higher-order relations that emerge from the interplay of object groups and open spaces, which entity- or pairwise-level representations struggle to capture.
To this end, we leverage \emph{3D foundation models} developed for point cloud understanding~\cite{partfield,sonata,concerto,utonia}, which can extract rich, scene context-aware features without additional task-specific training.

\textbf{Trajectory Planning.}
Trajectory planning seeks a time-indexed sequence of states connecting start and goal configurations, optionally through waypoints.
Classical methods minimize hand-designed costs~\citep{astar,dstar} or use randomized search~\citep{rrt,prob_roadmaps}, while exemplary learning-based methods use diffusion models~\citep{diffuser,potential_diffusion_planning,traj_stitching} or reinforcement learning~\citep{active_neural_slam,goat} for robustness under noisy observations.
All of these assume explicit start, goal, and waypoint specifications, whereas analogical trajectory transfer requires finding semantically analogous trajectories in a new scene with \emph{no such specifications}.
More recent line of work tackles trajectory planning considering scene semantics, where the goal is to generate 3D scene-conditioned motion trajectories that are faithful to a user text prompt.
To this end, existing approaches leverage multi-modal LLMs~\citep{event_driven_story,llm3_robot_plan} to plan keyframe motions or scene graphs~\citep{sayplan,tamp_3d_scene_graph} to generate trajectories that respect object relations. 
We adapt these approaches to our task setup and perform comparisons in Sec.~\ref{sec:experiments}.

\vspace{-1em}
\section{Method}
\label{sec:method}
\begin{figure*}[t]
  \centering
    \includegraphics[width=\linewidth]{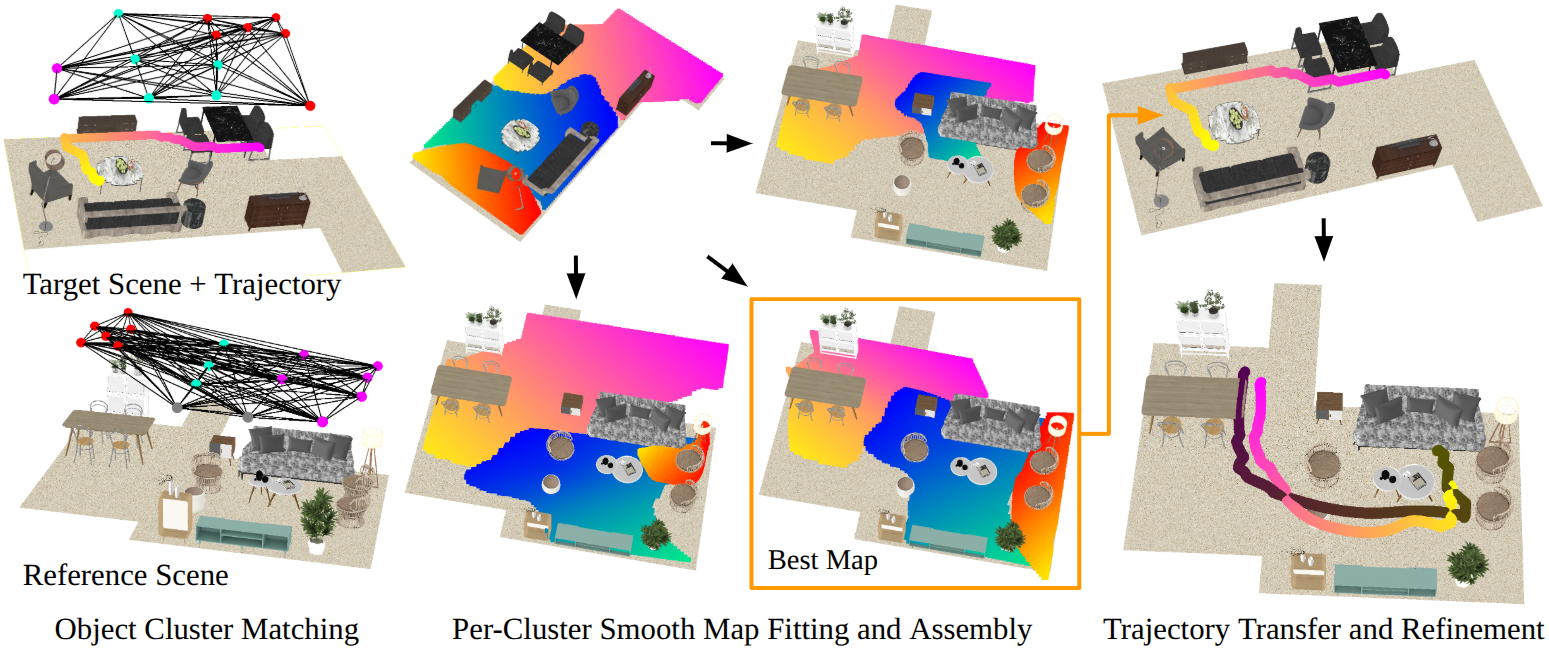}
   \caption{Overview of our approach. We first match object clusters (node colors represent cluster IDs) using object-level graphs and propagate them to the scene point clouds. For each cluster, we estimate a set of smooth maps and combinatorially assemble them into a scene-scale mapping. This mapping produces an initial trajectory transfer (black), which is further refined to avoid collisions and distortions (temporally color-coded).}
   \label{fig:overview}
   \vspace{-1em}
\end{figure*}
\subsection{Overview}
As shown in Fig.~\ref{fig:overview}, 
the inputs to our method are
 a target scene $\mathcal{S}_\text{tgt}$ and a reference scene $\mathcal{S}_\text{ref}$, each represented as a 3D point cloud with object instance labels without semantics.
In addition, a trajectory $\mathbf{T} = \{t_i\}_{i=1}^{M} \subset \mathbb{R}^3$ collected in $\mathcal{S}_\text{tgt}$ is given. 
Our goal is to produce a transferred trajectory $\hat{\mathbf{T}} \subset \mathcal{S}_\text{ref}$ that preserves the semantic structure of $\mathbf{T}$ while being spatially coherent and navigable in $\mathcal{S}_\text{ref}$.
To efficiently handle the large search space, our solution takes a hierarchical approach where scene-scale correspondences are first found for subdivided, low-granularity representations (object instance-level graphs), and adopted to high-granularity representations (smooth maps).
Specifically, our method partitions each scene into 
regions of object clusters
and finds matches between them (Sec.~\ref{sec:cluster_match}), fits a pool of candidate smooth maps per cluster pair (Sec.~\ref{sec:tps_fitting}), assembles them into a global scene map for initial transfer (Sec.~\ref{sec:tps_assembly}), and refines the transferred trajectory to minimize collisions and distortions (Sec.~\ref{sec:traj_transfer}).

\subsection{Object Cluster Matching and Region Propagation}
\label{sec:cluster_match}

\noindent\textbf{Object Graph Construction and Clustering.}
We extract per-point features from a pre-trained 3D foundation model~\citep{sonata,concerto} and represent each scene as an object graph $G = (V, E)$, where each node $v \in V$ stores the centroid and averaged features of one object instance, and edges connect nearby nodes and store the Euclidean distances between them.
Empirically, we find that Sonata ~\citep{sonata} features show the best overall performance.
We then apply Agglomerative Clustering~\citep{agglo_cluster} to the node coordinates, grouping objects into $C_\text{tgt}$ and $C_\text{ref}$ clusters for the two scenes.
Despite our simple clustering approach, the choice of clustering algorithm and hyperparameters does not largely affect performance (see Sec.~\ref{sec:supp_additional_ablation}).

\noindent\textbf{Graph-based Cluster Matching.}
To match clusters across scenes (Fig.~\ref{fig:overview}), we construct an affinity matrix $\mathbf{K} \in \mathbb{R}^{(N_\text{tgt} N_\text{ref}) \times (N_\text{tgt} N_\text{ref})}$, where $N_\text{tgt} = |V_\text{tgt}|$ and $N_\text{ref} = |V_\text{ref}|$.
The diagonal entry $K_{pq,pq}$ is the inner product of the foundation model feature vectors at target node $p$ and reference node $q$, while the off-diagonal entry $K_{pq,p'q'}$ encodes the geometric compatibility of simultaneously matching $(p,q)$ and $(p',q')$, computed as the inverse absolute difference of the corresponding edge lengths.
We pass $\mathbf{K}$ to a graph matching algorithm~\citep{rrvm_graph_match} to obtain a soft node assignment $\mathbf{X}_\text{assign} \in \mathbb{R}^{N_\text{tgt} \times N_\text{ref}}$, where $\mathbf{X}_\text{assign}[p, q]$ reflects the likelihood of matching target node $p$ to reference node $q$.
We then aggregate these into an inter-cluster matrix $\mathbf{X}_\text{inter} \in \mathbb{R}^{C_\text{tgt} \times C_\text{ref}}$, where entry $(i,j)$ captures how well the objects in target cluster $i$ collectively match those in reference cluster $j$, defined as $\mathbf{X}_\text{inter}[i, j] = \max_{\mathcal{M} \,\in\, \Pi(\mathcal{C}^i_\text{tgt},\, \mathcal{C}^j_\text{ref})} \sum_{(p, q) \in \mathcal{M}} \mathbf{X}_\text{assign}[p, q]$, with $\Pi(\mathcal{C}^i_\text{tgt}, \mathcal{C}^j_\text{ref})$ denoting all injections from cluster $i$ nodes $\mathcal{C}^i_\text{tgt}$ to cluster $j$ nodes $\mathcal{C}^j_\text{ref}$.
We apply a variant of the Hungarian algorithm~\citep{hungarian_topk} to $\mathbf{X}_\text{inter}$ to find the top-$K$ cluster assignments.
Here, low-confidence cluster matches (i.e., $\mathbf{X}_\text{inter}[i, j]$ below a threshold) are merged into their nearest valid cluster.

\noindent\textbf{Region Propagation.}
Cluster labels are then propagated to the full point cloud: object instance points inherit their instance's cluster ID, while open-space points inherit the nearest cluster centroid's ID, yielding a dense per-point labeling over the entire scene.

\subsection{Per-Cluster Smooth Map Fitting}
\label{sec:tps_fitting}

For each matched cluster pair, we estimate a set of candidate smooth maps 
modeled as a combination of similarity transforms and non-rigid deformations via thin-plate splines (TPS)~\citep{tps_1}.
To generate a diverse pool of candidates, we first enumerate rigid transformations by combining $N_r$ evenly-spaced y-axis rotations, four axis-aligned reflections, and translations derived from object centroid pairs in the two clusters.
We additionally consider per-axis scaling using the bounding box extent ratios of the cluster pair.
For each 
enumerated transformation, a smooth map is found by (i) applying the transformation to the target cluster points, (ii) assigning each target point a reference scene nearest neighbor, and (iii) fitting a TPS through these points $\phi(\cdot): \mathbb{R}^3 \rightarrow \mathbb{R}^3$ that maps target cluster points to the reference scene.
Candidate smooth maps are then scored by the mean feature discrepancy: for each target node $p$, we locate its spatially nearest neighbor $\texttt{NN}_\text{ref}(p)$ in the reference cluster after warping and measure the feature difference:
\begin{equation}
    \mathcal{L}_\text{feat}(\phi) = \frac{1}{|\mathcal{C}_\text{tgt}|} \sum_{p \in \mathcal{C}_\text{tgt}} \bigl\| f_\text{tgt}(p) - f_\text{ref}(\texttt{NN}_\text{ref}(p)) \bigr\|, \quad \texttt{NN}_\text{ref}(p) = \operatorname*{argmin}_{q \in \mathcal{C}_\text{ref}} \|\phi(p) - q\|,
    \label{eq:feat_cost}
\end{equation}
where $f_\text{tgt}(p)$ and $f_\text{ref}(q)$ denote the 3D foundation model features at points $p$ and $q$, respectively.
The top-$M$ maps with the smallest feature discrepancy are retained for each cluster pair.

\subsection{Per-Cluster Smooth Map Assembly}
\label{sec:tps_assembly}
Selecting the best smooth map for each cluster independently, without considering spatial and semantic context, yields globally inconsistent trajectory transfers.
We thus search for the optimal joint assignment of cluster-level smooth maps that best preserve the semantics while having small distortions and potential collisions.
Combining the results from the previous steps, we have a set of at most $K \times M$ (cluster match, smooth map) pairs for each of the $\mathcal{C}_\text{tgt}$ target scene clusters.
From here, we select one map per cluster to assemble a global scene map $\phi_\text{global}(\cdot)$.

We perform this selection by solving a combinatorial optimization problem whose goal is to minimize the total cost of an assignment $\mathcal{A}=[a_1, \dots a_{\mathcal{C}_\text{tgt}}]$, where $a_k$ denotes the index of the smooth map among the $K \times M$ maps per cluster.
To elaborate, we consider the sum of pairwise distortion costs and per-cluster navigability and feature costs, namely
\begin{equation}
    \mathcal{L}(\mathcal{A}) =
    \lambda_\text{feat} \sum_{k} \mathcal{L}_\text{feat}(\phi_{a_k}) +
    \lambda_\text{distort} \!\!\sum_{k < k'} \mathcal{L}_\text{distort}(a_k, a_{k'}) +
    \lambda_\text{nav} \sum_{k} \mathcal{L}_\text{nav}(\phi_{a_k}, k),
    \label{eq:assign_cost}
\end{equation}
where $\mathcal{L}_\text{feat}(\cdot)$ is identically defined as Eq.~\ref{eq:feat_cost},
For any two clusters $k$ and $k'$ with candidate maps $\phi_{a_k}$ and $\phi_{a_{k'}}$, the pairwise distortion cost measures the mean absolute change in inter-cluster distances:
\begin{equation}
    \mathcal{L}_\text{distort}(a_k, a_{k'}) = \frac{1}{N} \sum_{i=1}^{N} \bigl| \|\phi_{a_k}(\mathbf{p}_i) - \phi_{a_{k'}}(\mathbf{q}_i)\| - \|\mathbf{p}_i - \mathbf{q}_i\| \bigr|,
\end{equation}
where $\{\mathbf{p}_i\}$ and $\{\mathbf{q}_i\}$ are randomly sampled point pairs from clusters $k$ and $k'$, respectively.
In addition, the per-cluster navigability cost measures the fraction of trajectory points near each cluster that fall outside the navigable region of the reference scene after warping:
\begin{equation}
    \mathcal{L}_\text{nav}(\phi_{a_k}, k) = 1 - \frac{1}{|T_k|} \sum_{t \in T_k} \mathbf{1}\!\left[\min_{n \in \mathcal{N}} \|\phi_{a_k}(t)_{xz} - n\| \leq \delta\right],
    \label{eq:l_nav}
\end{equation}
where $T_k$ denotes the trajectory points assigned to cluster $k$, $\mathcal{N}$ is the set of navigable XZ positions in the reference scene, and $\delta$ is a proximity threshold.
Here navigable points $\mathcal{N}$ are defined as the open-space points of the reference scene, i.e., all points not belonging to any object instance annotation.
Finally, the total assignment cost is optimized via beam search~\citep{beam_search}, maintaining a beam of $L$ lowest-cost partial assignments across clusters.

\paragraph{Per-Cluster Smooth Map Merging}
From the top-1 beam solution, we concatenate the (target cluster points, smooth map-warped points) pairs of all selected per-cluster smooth maps and fit a single \emph{global
scene map} $\phi_\text{global}: \mathbb{R}^3 \to \mathbb{R}^3$.
This produces one smooth mapping covering the full scene, avoiding hard boundaries between cluster regions.
Note however, our method could also produce trajectory transfers using the top-$L$ beam solutions to support multi-modal applications as discussed in Sec.~\ref{sec:ablation}: we focus on using the top-1 predictions for performance benchmarking.

\subsection{Trajectory Transfer with Refinement}
\label{sec:traj_transfer}
Based on the global scene map $\phi_\text{global}$, we propose two transfer modes for mapping target scene trajectories to the reference scene.
Both transfer modes below apply $\phi_\text{global}$ to map locations from $\mathcal{S}_\text{tgt}$ to $\mathcal{S}_\text{ref}$; they differ in what is warped and how the final trajectory is assembled.

\noindent\textbf{Dense Trajectory Transfer.}
The first option is to warp the full source trajectory $\mathbf{T}$ through the assembled global map to produce an initial transferred trajectory.
This option focuses on preserving the fine-grained temporal structure of the original trajectory.
To remove remaining collision and distortion artifacts while preserving semantics, we refine the trajectory via gradient descent~\citep{sgd, adam} on the following cost:
\begin{equation}
    E = \lambda_\text{shape}\, E_\text{shape} + \lambda_\text{anchor}\, E_\text{anchor} + \lambda_\text{nav}\, E_\text{nav} + \lambda_\text{feat}\, E_\text{feat}.
    \label{eq:traj_warp_cost}
\end{equation}
The first three terms jointly remove collision and distortion artifacts: $E_\text{shape}$ preserves the original edge lengths, $E_\text{anchor}$ keeps points close to their initial warped positions $u_i = \phi_\text{global}(t_i)$, and $E_\text{nav}$ penalizes points outside the navigable region:
\begin{align}
    E_\text{shape}  &= \frac{1}{M-1} \sum_{i=1}^{M-1} \bigl(\|v_{i+1} - v_i\| - l_i\bigr)^2, \quad E_\text{anchor} = \frac{1}{M} \sum_{i=1}^{M} \|v_i - u_i\|, \\
    E_\text{nav}    &= \frac{1}{M} \sum_{i=1}^{M} \max\!\bigl(0,\; \tau - \rho_i\bigr)^2, \quad \rho_i = \sum_{n \in \mathcal{N}} \exp\!\Bigl(-\tfrac{\|v_i - n\|^2}{2\sigma^2}\Bigr),
\end{align}
where $l_i$ are the original edge lengths, $\rho_i$ is the Gaussian kernel density estimate of navigable points at $v_i$ with bandwidth $\sigma$, $\tau$ is a minimum density threshold, and $\mathcal{N}$ are navigable points as in Eq.~\ref{eq:l_nav}.
Intuitively, $\rho_i$ measures how surrounded $v_i$ is by navigable points, so the cost penalizes trajectory points that lie far from any navigable region.
The last term $E_\text{feat} = \frac{1}{|\mathbf{T_\text{S}}|} \sum_{i \in \mathbf{T_\text{S}}} \|v_i - \tilde{v}_i\|$ preserves the semantic structure between trajectories by attracting a sparse set $\mathbf{T_\text{S}} \subset \mathbf{T}$ of sampled trajectory points toward feature-matched positions $\tilde{v}_i$ in the reference scene, where $\tilde{v}_i$ is a point from the spatial neighborhood of the initial warp $u_i$ whose 3D foundation model features best match that of the original trajectory point $t_i$.

\noindent\textbf{Waypoint Transfer with Classical Planning.}
As an alternative, our method supports waypoint-based transfer, 
where only a sparse sequence of $N$ waypoints $\mathcal{W} = \{w_i\}_{i=1}^{N} \subset \mathbf{T}$
of the source trajectory are warped and refined using the same objective (Eq.~\ref{eq:traj_warp_cost}).
Note that the refinement procedure is still necessary since initial waypoint transfers may lie in non-traversable areas such as object interiors.
After this, we apply classical A* planning~\citep{astar} between consecutive waypoint pairs $(\hat{w}_i, \hat{w}_{i+1})$ in the reference scene to produce a transferred trajectory.
Compared to dense trajectory transfer, this mode ensures that the path between waypoints is collision-free as long as classical planning succeeds, at the cost of slightly sacrificing fine-grained trajectory structure preservation.

\section{Experiments}
\label{sec:experiments}
\begin{table*}[t]
\centering
\caption{
  Results on 3D-FRONT and ARKitScenes with paired single-GT and single-side annotations.
  \colorbox{red!30}{\strut red}, \colorbox{orange!40}{\strut orange}, and \colorbox{yellow!50}{\strut yellow} cells indicate the 1st, 2nd, and 3rd best result per column, respectively.
}
\label{tab:eval_single_auto}
\resizebox{\textwidth}{!}{%
\setlength{\tabcolsep}{5pt}
\begin{tabular}{l cc ccc| cc cc}
\toprule
& \multicolumn{2}{c}{3D-FRONT Paired} & \multicolumn{3}{c}{3D-FRONT Single-Side} & \multicolumn{2}{c}{ARKit Paired} & \multicolumn{2}{c}{ARKit Single-Side} \\
\cmidrule(lr){2-3}\cmidrule(lr){4-6}\cmidrule(lr){7-8}\cmidrule(lr){9-10}
Method
  & \makecell{Traj.\\AED\,$\downarrow$}
  & \makecell{Inlier\\@1.5\,$\uparrow$}
  & \makecell{Coll.\\Ratio\,$\downarrow$}
  & \makecell{Feat.\\Dist.\,$\downarrow$}
  & \makecell{Runtime\\(s)\,$\downarrow$}
  & \makecell{Traj.\\AED\,$\downarrow$}
  & \makecell{Inlier\\@1.5\,$\uparrow$}
  & \makecell{Coll.\\Ratio\,$\downarrow$}
  & \makecell{Feat.\\Dist.\,$\downarrow$} \\
\midrule
Feature NN                            & 1.597 & 0.474 & 0.227 & \rankone{0.820} & \rankone{0.20} & 1.493 & 0.537 & 0.072 & 0.928 \\
Neural Contextual Scene Maps~\cite{}  & 1.726 & 0.494 & \rankone{0.033} & 0.906 & 2.31 & 1.563 & 0.457 & \rankone{0.012} & 0.947 \\
Foundation Model Analogies            & 1.866 & 0.408 & 0.061 & 0.967 & 8.91 & 1.617 & 0.532 & \ranktwo{0.016} & 0.962 \\
Monte Carlo Localization              & 2.057 & 0.406 & 0.055 & 0.874 & 1.87 & 1.722 & 0.457 & 0.033 & \rankthree{0.921} \\
Object Node Graph Match~\cite{}       & \rankthree{1.475} & 0.525 & 0.216 & 0.965 & \ranktwo{0.32} & 1.567 & 0.499 & 0.109 & 0.969 \\
Object Node Graph Match + ICP~\cite{} & 1.512 & \rankthree{0.556} & 0.248 & 0.955 & 0.45 & 1.461 & \rankthree{0.591} & 0.149 & 0.977 \\
Keypoints Graph Match                 & 1.883 & 0.404 & 0.196 & 0.955 & \rankthree{0.35} & \rankthree{1.435} & 0.568 & 0.133 & 0.981 \\
Multi-modal LLM (Dense)~\cite{}       & 1.869 & 0.432 & 0.181 & 0.969 & 2.13 & 1.473 & 0.549 & 0.132 & 0.962 \\
Multi-modal LLM (Waypoint)~\cite{}    & 1.715 & 0.473 & 0.229 & 0.990 & 2.70 & 1.869 & 0.432 & 0.130 & 0.961 \\
\midrule
Ours (Dense)                          & \rankone{0.997} & \rankone{0.672} & \ranktwo{0.037} & \ranktwo{0.821} & 0.66 & \rankone{1.088} & \ranktwo{0.664} & \ranktwo{0.016} & \rankone{0.919} \\
Ours (Waypoint)                       & \ranktwo{1.070} & \ranktwo{0.645} & \rankthree{0.049} & \rankthree{0.867} & 0.61 & \ranktwo{1.126} & \rankone{0.680} & \rankthree{0.019} & \rankone{0.919} \\
\bottomrule
\end{tabular}%
}
\end{table*}

We evaluate analogical trajectory transfer on both synthetic and real-world datasets (Sec.~\ref{sec:traj_eval}, Sec~\ref{sec:ablation}), and show applications in a wide range of downstream tasks (Sec.~\ref{sec:applications}).

\noindent\textbf{Datasets.}
\emph{3D-FRONT}~\citep{td_front} is a large-scale dataset of synthetic indoor scenes with diverse room types and object layouts.
\emph{ARKitScenes}~\citep{arkit} consists of real-world 3D scans of indoor spaces captured with consumer-grade depth cameras, exhibiting sensor noise, missing regions, and partial object observations.

\noindent\textbf{Annotations.}
As our task is new, we design three types of trajectory annotations for evaluation.
\emph{Single-side} annotations are constructed by randomly sampling waypoints in the \emph{target scene only} and connecting them with A* planning; thus this annotation type could be collected automatically for a large number of scene pairs.
\emph{Paired (single GT)} annotations are collected by having human annotators select waypoints in scene pairs such that the resulting A*-planned trajectories are contextually similar; each pair has one ground-truth reference trajectory.
\emph{Paired (multiple GT)} annotations extend the single-GT setting by having multiple annotators independently label the same scene pair, yielding multiple ground-truth reference trajectories that reflect the inherent multi-modality of valid trajectory transfers.

\noindent\textbf{Metrics.}
For the \emph{Single-side} annotations, we report i) \emph{Collision ratio}: fraction of transferred trajectory points within a distance threshold of obstacle points; ii) \emph{Feature distance}: mean 3D foundation model feature discrepancy between source and transferred trajectory points, where scene features are interpolated to trajectory point locations as in~\citep{sparsedff}.
Next, for the \emph{Paired (single, multiple GT)} annotations, we report i) \emph{Trajectory AED}: median Euclidean distance between the predicted and ground-truth trajectories, and ii) \emph{Inlier ratio}: fraction of predicted trajectory points within an inlier distance threshold of their ground-truth counterparts.
These metrics are inspired from prior literature in visual localization~\citep{locnerf,mapfreeloc}.
For the multiple-GT setting, all GT-based metrics are reported as the best value across annotated ground-truth samples, reflecting that any valid transfer should be rewarded.

\subsection{Trajectory Transfer Evaluation}
\label{sec:traj_eval}
\noindent\textbf{Baselines.}
We compare our method against adaptations of 3D scene understanding pipelines to our task.
First, the \emph{Feature NN} baseline retrieves the nearest neighbor in the reference scene via 3D foundation model~\citep{sonata} feature distance, and connects the matched waypoints with A* planning to form the transferred trajectory.
Second, \emph{Neural Contextual Scene Maps}~\citep{scene_analogies} and \emph{Foundation Model Analogies}~\citep{fma} find a \emph{single} smooth map as opposed to ensembles in our method based on learned task-specific or 2D foundation model features, and use the smooth map to transfer trajectories.
Third, the \emph{Monte Carlo Localization}~\citep{monte_carlo_loc} baseline adapts particle filtering-based localization for our task by setting the particle weights proportional to 3D foundation model feature affinity, similar to Loc-NeRF ~\citep{locnerf}.
Fourth, the graph matching-based baselines i) build a graph with varying granularity, i.e., object-centric (\emph{Object Node Graph Match}) or keypoint-based (\emph{Keypoints Graph Match}), ii) match the graphs with 3D foundation model features assigned to each node along with an \emph{optional ICP}~\citep{icp} step to obtain fine-grained matches from object-level matches, and iii) transfer trajectories using a thin-plate spline fit to the matches.
Finally, the \emph{Multi-modal LLM}~\citep{gpt5} baselines are given top-down renderings of the scene pairs, with target scene waypoints or trajectory points overlaid, and are prompted to select contextually corresponding locations in the reference scene.

\setlength{\intextsep}{-3pt}
\begin{wraptable}{R}{0.5\columnwidth}
\centering
\caption{
  Results under the paired multiple ground-truth and cross-domain transfer settings.
}
\label{tab:multi_ref_sim2real}
\vspace{3pt}
\resizebox{0.5\columnwidth}{!}{%
\setlength{\tabcolsep}{4pt}
\begin{tabular}{lcc cc}
\toprule
& \multicolumn{2}{c}{Paired (Multiple GT)} & \multicolumn{2}{c}{Cross-Domain} \\
\cmidrule(lr){2-3}\cmidrule(lr){4-5}
Method
  & \makecell{Traj.\\AED\,$\downarrow$}
  & \makecell{Inlier\\@1.5\,$\uparrow$}
  & \makecell{Traj.\\AED\,$\downarrow$}
  & \makecell{Inlier\\@1.5\,$\uparrow$} \\
\midrule
Feature NN               & 1.770 & 0.408 & 1.858 & 0.399 \\
Fnd. Model Analogies            & 1.413 & 0.551 & 1.668 & 0.438 \\
Monte Carlo Loc.                      & \rankthree{1.196} & \rankthree{0.616} & 1.715 & 0.423 \\
Obj. Graph Match+ICP~\cite{}         & 1.307 & 0.578 & \ranktwo{1.528} & \rankthree{0.441} \\
MLLM (Waypoint)~\cite{}  & 1.447 & 0.585 & 2.082 & 0.326 \\
\midrule
Ours (Dense)             & \rankone{0.879} & \ranktwo{0.676} & \rankthree{1.577} & \ranktwo{0.522} \\
Ours (Waypoint)          & \ranktwo{0.958} & \rankone{0.677} & \rankone{1.464} & \rankone{0.570} \\
\bottomrule
\end{tabular}%
}
\end{wraptable}

\begin{figure*}[t]
  \centering
    \includegraphics[width=\linewidth]{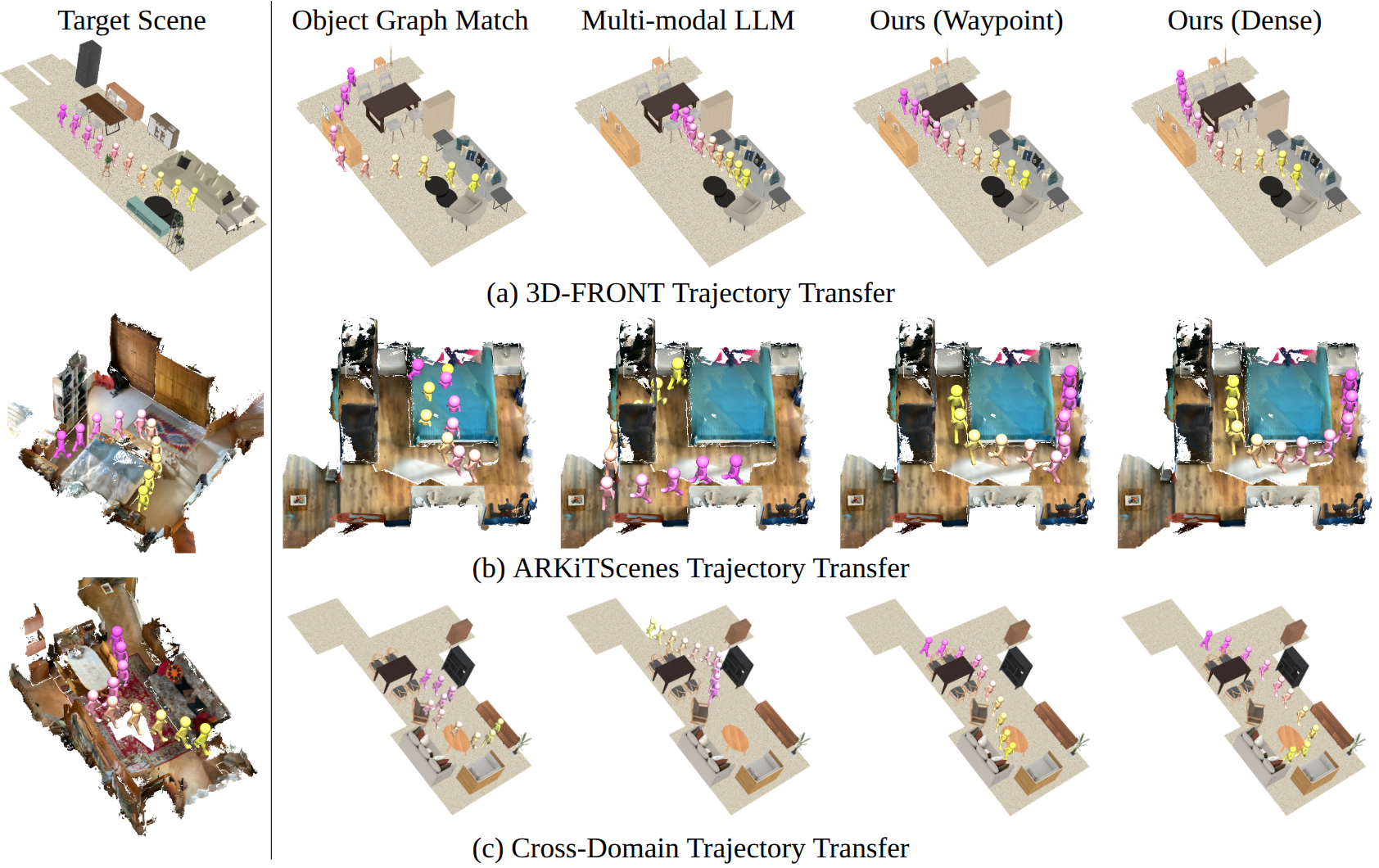}
   \caption{Qualitative comparison of analogical trajectory transfer on various datasets and baselines.}
   \label{fig:qualitative}
\end{figure*}

\noindent\textbf{Performance Comparison.}
Tables~\ref{tab:eval_single_auto} and~\ref{tab:multi_ref_sim2real} show the quantitative evaluation results in 3D-FRONT and ARKiTScenes datasets, while qualitative comparisons are shown in Fig.~\ref{fig:qualitative}.
For 3D-FRONT scenes, our method outperforms all baselines across GT-free and GT-based metrics, demonstrating its ability to produce semantically consistent transfers in diverse synthetic layouts with minimal collision.
Our method further maintains strong performance in ARKitScenes under real-world scan noise and missing geometry.
By using 3D foundation model features, our method can stay robust in both scenarios.
In terms of runtime, our method can run in around 0.65 sec. using an A6000 GPU, which is shorter than baselines that employ costly LLMs while showing accurate transfer results.

\setlength{\intextsep}{-2pt}
\begin{wraptable}{R}{0.5\columnwidth}
\centering\footnotesize
\caption{
  Ablation study results.
  We report inlier ratios ($\uparrow$) at various thresholds.
}
\label{tab:ablation_main}
\vspace{3pt}
\resizebox{0.5\columnwidth}{!}{%
\setlength{\tabcolsep}{4pt}
\begin{tabular}{lcccc}
\toprule
Method & @0.75\,$\uparrow$ & @1.00\,$\uparrow$ & @1.25\,$\uparrow$ & @1.50\,$\uparrow$ \\
\midrule
Ours w/ CLIP          & 0.251 & 0.322 & 0.422 & 0.505 \\
Ours w/ PartField     & 0.270 & 0.349 & 0.420 & 0.482 \\
Ours w/o Obj.\ Graph  & 0.324 & 0.432 & 0.516 & 0.608 \\
Ours w/o Traj.\ Opt.  & 0.373 & 0.473 & 0.570 & 0.664 \\
\midrule
Ours (Dense)          & 0.398 & 0.514 & 0.601 & 0.677 \\
Ours Top-3            & 0.559 & 0.681 & 0.736 & 0.770 \\
Ours Top-5            & 0.581 & 0.699 & 0.763 & 0.822 \\
Ours Top-10           & \textbf{0.640} & \textbf{0.739} & \textbf{0.800} & \textbf{0.847} \\
\bottomrule
\end{tabular}%
}
\end{wraptable}

The \emph{Feature NN}, \emph{Monte Carlo Localization}, and \emph{Graph Matching} baselines underperform in both regimes, which indicates that naive feature-space search or directly matching graphs without object clustering incurs spatially incoherent mappings (e.g., dining table chairs being mapped to distant locations in the reference scenes), leading to sub-optimal performance.
Baselines that produce a scene-scale smooth map \emph{single-shot} also show lower performance (\emph{Neural Contextual Scene Maps}, \emph{Foundation Model Analogies}), since fitting a single map to cover the entire scene has an immense search space.
Our method, in contrast, adopts a divide-and-conquer strategy, which enables managing the large search space and producing stable transfer results.
Finally, the \emph{multi-modal LLM} baselines also struggle to accurately transfer trajectory points despite the impressive performance of LLMs in vision-language benchmarks~\citep{spatial_vlm,3dqa}.
Nevertheless, we believe that multi-modal LLMs could perform better in our task through instruction fine-tuning, yet developing prompting strategies and pipelines for collecting a large number of training annotations remains an open area for research.

\noindent\textbf{Cross-Domain Transfer.}
Table~\ref{tab:multi_ref_sim2real} and Fig.~\ref{fig:qualitative} report results on cross-domain (3D-FRONT $\leftrightarrow$ ARKitScenes) scene pairs, evaluating generalization across the synthetic-to-real domain gap.
Our method outperforms all baselines, demonstrating that 3D foundation model features provide robust cross-domain scene context.

\subsection{Ablation Study}

\label{sec:ablation}
\begin{wrapfigure}{R}{0.4\columnwidth}
  \centering
  \includegraphics[width=0.4\columnwidth]{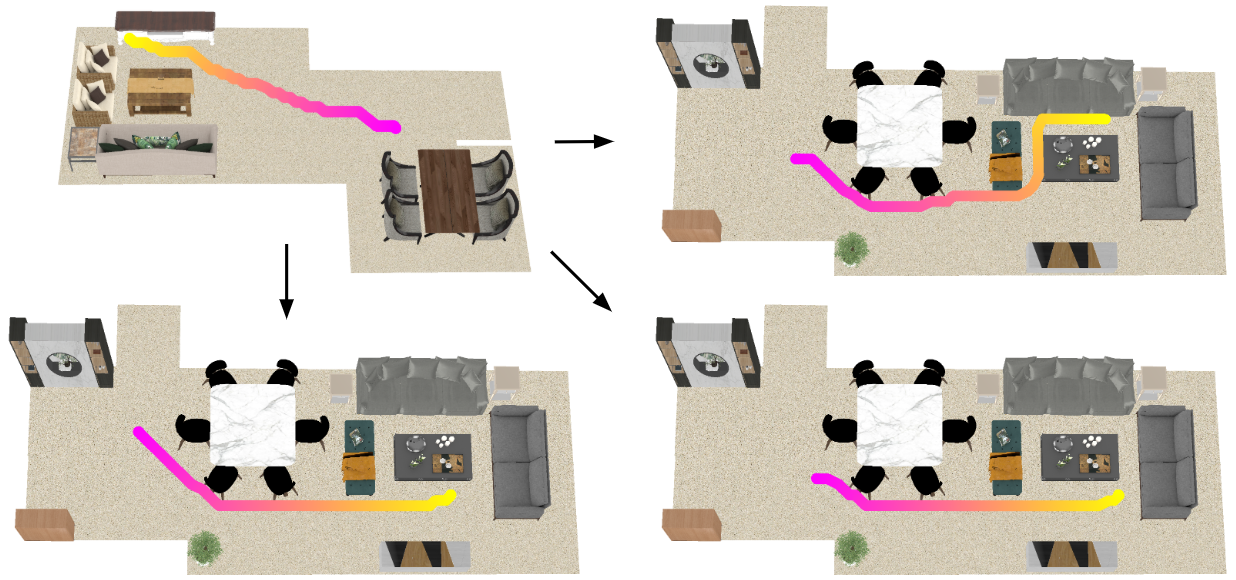}
  \caption{Visualization of top-$L$ trajectory transfer candidates in 3D-FRONT.}
  \label{fig:multi_modal}
\end{wrapfigure}

Table~\ref{tab:ablation_main} and Fig.~\ref{fig:multi_modal} report ablation results on paired (multiple GT) annotations from 3D-FRONT and ARKiTScenes.
Replacing 3D Foundation Model features that handle scene-scale point clouds~\citep{sonata} with 2D (CLIP~\citep{clip}) or 3D Foundation Models that focus on object-level point clouds (PartField~\citep{partfield}) degrades performance substantially, confirming that scene-level features are critical for accurate cluster matching and smooth map scoring.
Removing the object graph during cluster matching (\emph{w/o Obj.\ Graph}) hurts performance as well, showing that consideration of object-level relationships is important for preventing spatially incoherent maps.
Further, removing the trajectory optimization step (\emph{w/o Traj.\ Opt.}) also degrades performance, demonstrating the importance of refining the initial warp to remove residual collisions and distortions.
Finally, we evaluate whether our method can produce multi-modal predictions, which is critical as our task is inherently multi-modal and admits multiple valid solutions.
Reporting the oracle best among Top-$K$ candidates from Sec.~\ref{sec:tps_assembly} (instead of Top-1) yields a large improvement, confirming that our method could output multiple reasonable transfers as shown in Fig.~\ref{fig:multi_modal}.

\subsection{Applications}
\label{sec:applications}
\begin{figure*}[t]
  \centering
    \includegraphics[width=\linewidth]{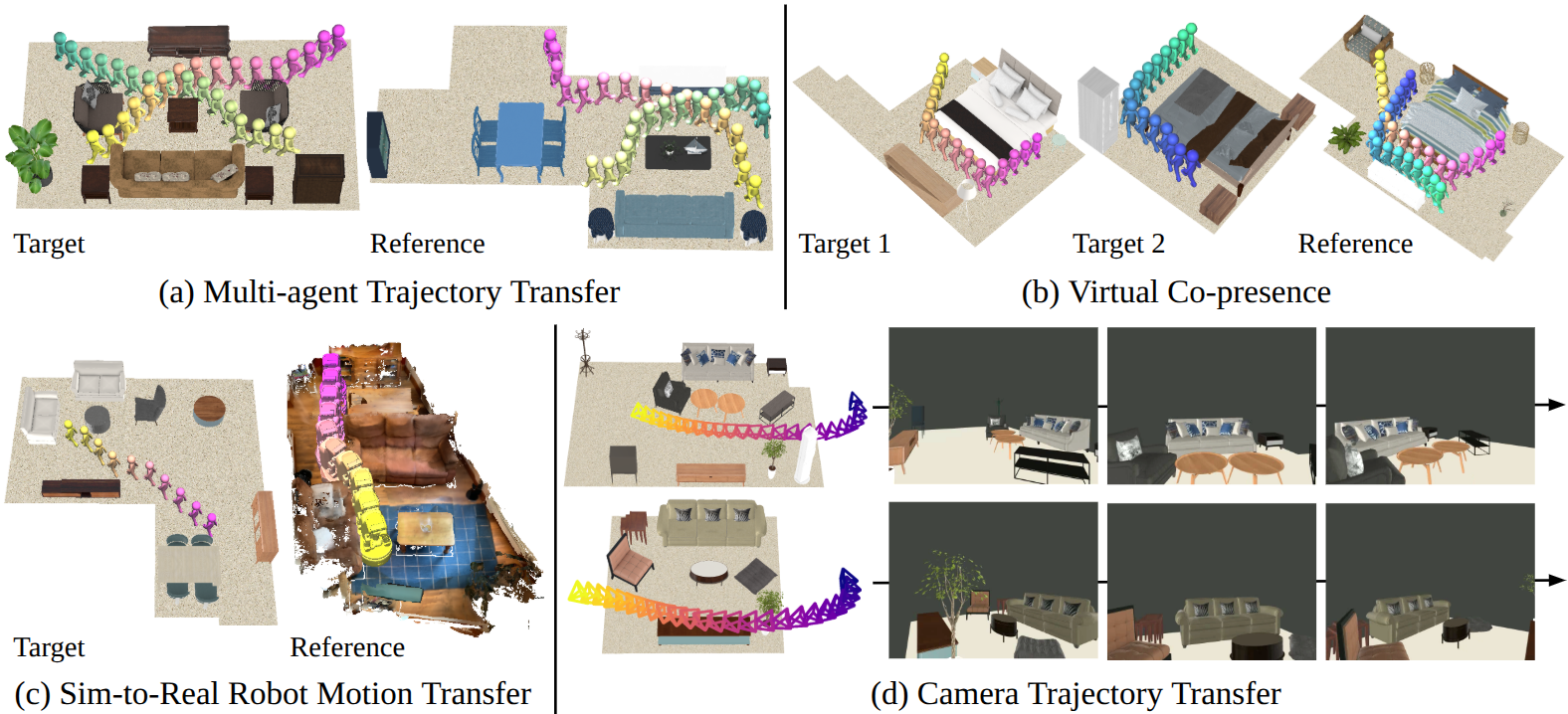}
   \caption{Visualizations of applications enabled by analogical trajectory transfer.}
   \label{fig:applications}
\end{figure*}

In Fig.~\ref{fig:applications}, we show applications of analogical trajectory transfer to various downstream tasks in AR/VR and robotics.

\noindent\textbf{Multi-agent Trajectory Transfer.}
Our method can simultaneously transfer trajectories of multiple agents while preserving their relative interactions.
This enables direct applications in AR/VR content creation, where multiple participants' activities need to be re-staged in a new environment.

\setlength{\intextsep}{-2pt}

\noindent\textbf{Sim-to-Real Robot Motion Transfer.}
Human motion demonstrations collected in synthetic environments can be transferred to robot trajectories in real-world scans using our method, bridging the domain gap in data collection for mobile robot training.
Backed by advances in human motion generation from synthetic 3D scenes~\citep{trumans}, our method could aid in building scalable pipelines for generating mobile robot training data.

\noindent\textbf{Virtual Co-presence.}
Given trajectories in multiple different scenes, our method can transfer all of them into a \emph{single} shared reference scene.
This enables virtual co-presence scenarios where participants from separate physical environments aim to share a common virtual space~\citep{copresence}.

\noindent\textbf{Camera Trajectory Transfer.}
Our method can transfer camera scan paths, producing videos that capture analogous contextual information across environments.
Potential applications include automated scan path planning and house tour video creation~\citep{housetour}.

\section{Conclusion}
\label{sec:conclusion}
We introduced \textit{analogical trajectory transfer}, the task of translating trajectory points in one scene to semantically analogous locations in another.
To cope with the wide variability of scene arrangements and large search space, we proposed a training-free pipeline that hierarchically predicts smooth maps using 3D foundation model features.
Our method partitions scenes into object-centric clusters using 3D foundation model features, assembles per-cluster smooth maps into a global scene map, and refines the transferred trajectory via gradient descent.
Experiments show that our method can robustly find physically plausible trajectory transfers on synthetic and real-world scenes by effectively exploiting 3D foundation model features.
We further demonstrated the applicability of our method to tasks such as sim-to-real robot motion transfer and virtual co-presence, which highlights the wide range of downstream use cases enabled by our approach.
We expect our method to serve as a foundation for future work on analogical spatial reasoning, enabling broader research into how human and robot activities can be transferred and adapted across diverse 3D environments.

\noindent\textbf{Limitations.}
Our method currently operates on static scene pairs: when agents interact with objects and alter the environment, an update mechanism that respects object dynamics would be needed for reactive planning.
Our method also targets scene pairs that share similar semantics and functionality, and extending it to transfers across largely different scene types (e.g., bedroom to living room) using richer cues such as affordance features remains an open challenge.
On the evaluation side, our paired annotations rely on human annotators while single-side evaluations use only proxy metrics; a more scalable protocol possibly based on procedural scene generation would strengthen future benchmarking.
Finally, applying our method to other scene types such as tabletop or outdoor environments is a natural direction for future work.


{
\small

\bibliographystyle{plainnat}
\bibliography{main}

}


\appendix
\section{Hyperparameters}
\label{sec:supp_hyperparams}

We summarize key hyperparameters used in our method below.

\noindent\textbf{Point Cloud Preprocessing.}
Input point clouds are voxelized into a grid of size $0.02$\,m before extracting Sonata~\citep{sonata} features.
Object instance nodes from Sec.~\ref{sec:cluster_match} are connected by a fully-connected graph (edge distance threshold set to a large value), with edge weights storing pairwise Euclidean distances.

\noindent\textbf{Graph Matching.}
Cluster matching uses top-$K=1$ graph-level matches per target cluster for 3D-FRONT~\citep{td_front} and $K=3$ for ARKitScenes~\citep{arkit}.
For Agglomerative Clustering, we set the target cluster size to be $4$ objects per cluster.
This results in cluster sizes that are around $4$.

\noindent\textbf{Per-Cluster Smooth Map Fitting.}
For 3D-FRONT, we generate $N_r = 4$ evenly-spaced y-axis rotations combined with four axis-aligned reflections, yielding up to $16$ rigid initialization candidates per object centroid pair.
For ARKitScenes, we set $N_r = 8$ and consider the same number of axis-aligned reflections.
Then, the top-$M = 5$ smooth maps with the lowest feature discrepancy are retained per cluster pair.

\noindent\textbf{Smooth Map Assembly.}
The beam search width in Sec.~\ref{sec:tps_assembly} is $L = 5$.
The cost weights for the assembly objective (Eq.~\ref{eq:assign_cost}) are:
$\lambda_\text{feat} = 1.0$, $\lambda_\text{distort} = 1.0$, $\lambda_\text{nav} = 1.0$.

\noindent\textbf{Trajectory Refinement.}
The cost weights for the trajectory refinement objective (Eq.~\ref{eq:traj_warp_cost}) are:
$\lambda_\text{shape} = 1.0$, $\lambda_\text{anchor} = 0.1$, $\lambda_\text{nav} = 1.0$, $\lambda_\text{feat} = 1.0$.
The navigable density bandwidth is $\sigma = 0.2$\,m, and the minimum density threshold is set to the average density field value for the navigable points $\mathcal{N}$.

\section{Baseline Implementations}
\label{sec:supp_baselines}

We describe how each baseline is adapted to the trajectory transfer task.

\noindent\textbf{Feature NN.}
For each target scene waypoint, we retrieve its nearest neighbor in the reference scene by Euclidean distance in the Sonata~\citep{sonata} feature space.
The matched reference positions connected via A* planning, to produce a dense trajectory transfer.

\noindent\textbf{Neural Contextual Scene Maps~\citep{scene_analogies}.}
The original method learns a neural feature field that can describe the object relationships for each scene and finds a smooth map between user-specified source and target regions, parameterized as an affine transform plus a local deformation.
We adapt it to trajectory transfer by setting the bounding region of the source trajectory as the whole input region and running the method to obtain a smooth map, which is then applied to the trajectory points.

\noindent\textbf{Foundation Model Analogies~\citep{fma}.}
This method builds a object-level graph over each scene and equips each object node with CLIP~\citep{clip} features averaged across multi-view renderings of the object.
Then, the method applies graph matching to obtain a global smooth map by connecting object node-wise matches.
PartField~\citep{partfield} features are then used for local refinement, where the initial smooth map is optimized to minimize the PartField feature discrepancies between the source and target scene.
We adapt the method to trajectory transfer by applying the resulting smooth map to the trajectory points.

\noindent\textbf{Monte Carlo Localization.}
We adapt particle filtering-based localization~\citep{monte_carlo_loc} for trajectory transfer.
A set of particles is initialized uniformly over the reference scene, and particle weights are set proportional to the Sonata~\citep{sonata} feature affinity between each particle's location and the corresponding source trajectory point.
The sequence of highest-weight particle locations across time steps forms the transferred trajectory.

\noindent\textbf{Object Node Graph Match.}
We build an object-level scene graph for each scene, assigning each detected object instance a node whose feature is the mean Sonata~\citep{sonata} feature over the object's point cloud, and apply graph matching~\citep{rrvm_graph_match} to obtain object-level correspondences.
A thin-plate spline~\citep{tps_1} is fit to the matched object centroids, and the resulting smooth map transfers the trajectory.

\noindent\textbf{Object Node Graph Match + ICP.}
The same pipeline as Object Node Graph Match, with an additional ICP~\citep{icp} step applied after graph matching to rigidly align the matched object point sets before fitting the thin-plate spline.
ICP refines the initial graph-matching correspondences by minimizing point-to-point distances, providing a more accurate initialization for the smooth map.

\noindent\textbf{Keypoints Graph Match.}
Instead of object-level nodes, we build a keypoint graph using up to $1000$ sampled points from the scene point cloud, with Sonata~\citep{sonata} features assigned to each keypoint.
Graph matching, thin-plate spline fitting, and trajectory transfer follow the same procedure as Object Node Graph Match.

\noindent\textbf{Multi-modal LLM (Dense)~\citep{gpt5}.}
A multi-modal LLM (GPT-5.1) is given top-down renderings of the scene pair with all regularly sampled target scene trajectory points overlaid on the target scene rendering.
The model is prompted to output the corresponding reference-scene position for each sampled trajectory point, and the predicted trajectory points are connected using a Cubic Spline to produce a dense transferred trajectory without any planning step.

\noindent\textbf{Multi-modal LLM (Waypoint)~\citep{gpt5}.}
A multi-modal LLM (GPT-5.1) is given top-down renderings of the scene pair with the target scene waypoints overlaid on the target scene rendering.
The model is prompted to select the contextually corresponding waypoint locations in the reference scene, and the selected waypoints are then connected with A* planning~\citep{astar} to produce the transferred trajectory.

\section{Experiment Setup Details}
\label{sec:supp_exp_setup}

\noindent\textbf{Compute Resources.} We conduct all our experiments using RTX 3090 and A6000 GPUs.

\noindent\textbf{Annotation Counts.}
We collect 77, 85, and 27 paired (single GT) scene pairs for 3D-FRONT, ARKitScenes, and cross-domain, respectively.
The paired (multiple GT) split contains 24 and 22 scene pairs for 3D-FRONT and ARKitScenes, respectively.
The single-side split contains 201 scene pairs for each of 3D-FRONT and ARKitScenes.

\noindent\textbf{Single-Side Annotation.}
Single-side waypoints are obtained by applying farthest point sampling to the navigable space of the target scene to obtain a diverse set of candidate positions.
Consecutive sampled positions are connected with A* planning~\citep{astar} to form a valid, collision-free trajectory.
This procedure requires no human annotation and can be scaled to a large number of scene pairs automatically.

\noindent\textbf{Paired Annotation.}
Human annotators are presented with a pair of 3D scenes rendered from a top-down view via the annotation GUI described in Sec.~\ref{sec:supp_annot}.
Annotators select a sequence of waypoints in the target scene, then select contextually corresponding waypoints in the reference scene.
The selected waypoints in each scene are connected with A* planning to form the trajectory pair.
For multiple-GT annotations, multiple annotators independently label the same scene pair, yielding several valid ground-truth reference trajectories.
In our experiments, we asked 4 human annotators to label the scene pairs.

\noindent\textbf{Cross-Domain Pairs.}
Cross-domain pairs are formed by pairing 3D-FRONT synthetic scenes with ARKitScenes real-world scans that share similar room types.
These pairs test the method's ability to generalize across the sim-to-real domain gap.

\section{Trajectory Transfer Annotation Collection}
\label{sec:supp_annot}
\begin{figure*}[t]
  \centering
    \includegraphics[width=0.6\linewidth]{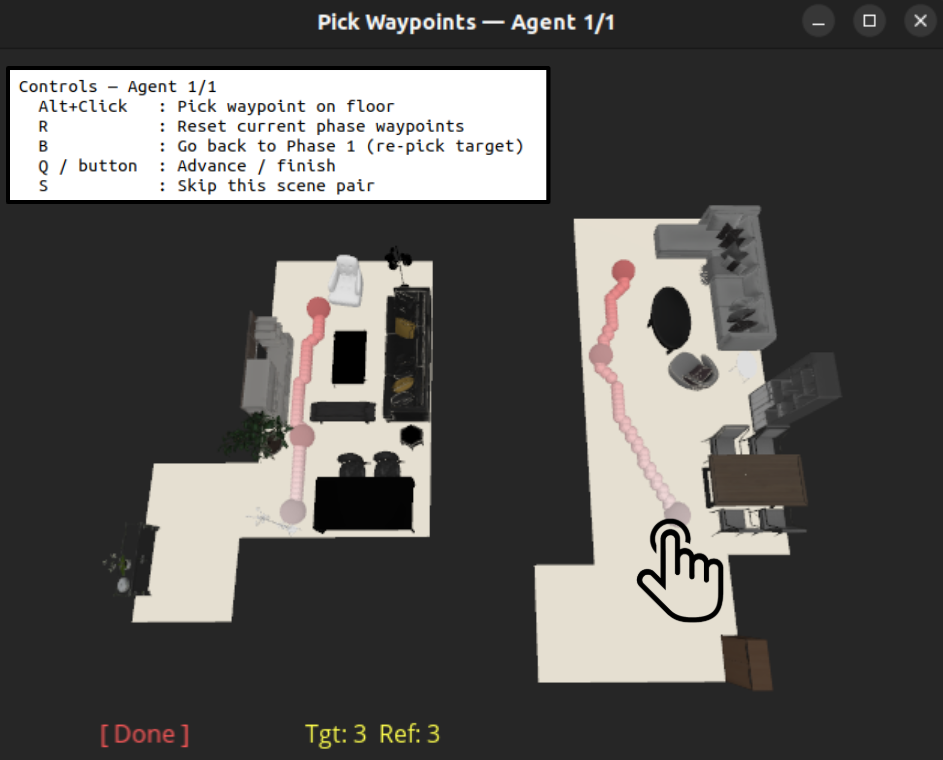}
   \caption{Graphical user interface (GUI) used for paired (single- and multiple ground-truth) annotations. Users are prompted to select corresponding waypoints from the pair of scenes displayed on screen, and the waypoints are connected to produce trajectory transfer pairs.}
   \label{fig:gui}
\end{figure*}

\noindent\textbf{Scene Pair Selection.}
Scene pairs are selected manually by inspecting top-down renderings of candidate scenes.
Annotators select pairs whose room types and functional layouts are similar enough to admit a meaningful trajectory transfer (e.g., two living rooms with comparable object arrangements).
Cross-domain pairs are formed by matching 3D-FRONT synthetic scenes with ARKitScenes real-world scans of the same room type.

\noindent\textbf{Annotation GUI.}
Annotations are collected using a custom GUI (Fig.~\ref{fig:gui}) that displays side-by-side top-down renderings of the target and reference scenes.
The annotator first places waypoints in the target scene by clicking on the top-down view, forming a trajectory that visits semantically meaningful locations.
The GUI then runs A* planning between consecutive waypoints to produce a collision-free trajectory in the target scene, which is displayed to the annotator for confirmation.
The annotator then places the corresponding waypoints in the reference scene such that the resulting planned trajectory visits analogous locations.

\noindent\textbf{Quality Control.}
After annotation, each trajectory pair is visually reviewed to ensure that the reference trajectory visits locations semantically corresponding to those in the target trajectory.
Pairs with ambiguous or low-quality annotations are discarded.
For multiple-GT annotations, four annotators independently label each scene pair, and all valid annotations are retained as ground-truth references.

\section{Additional Results}
\label{sec:supp_additional_results}

\subsection{Additional Metrics and Full Results}

In this section, we share the full quantitative evaluations on the 3D-FRONT~\citep{td_front} and ARKitScenes~\citep{arkit} datasets using additional metrics.

\noindent\textbf{Length Distortion.}
Length distortion measures the geometric distortion introduced by the transfer by comparing per-segment lengths.
Formally, it is the mean per-segment ratio $|l_\text{ref} - l_\text{src}| / l_\text{src}$, where $l_\text{src}$ and $l_\text{ref}$ are the lengths of corresponding segments in the source and transferred trajectories.
A lower value indicates that the transferred trajectory better preserves the original path shape.

\noindent\textbf{Waypoint AED.}
Waypoint AED is the median Euclidean distance between the transferred waypoints and their ground-truth counterparts in the reference scene.
This metric directly evaluates the accuracy of waypoint-level localization, complementing Trajectory AED which operates over densely resampled trajectories.

\noindent\textbf{Full Results.}
Tab.~\ref{tab:supp_full_results} reports full results on 3D-FRONT and ARKitScenes, including Waypoint AED and Inlier Ratio at two thresholds for the paired split, and Length Distortion for the single-side split.
Our method achieves the best Waypoint AED and Inlier Ratio across both datasets, while Length Distortion is lowest for Neural Contextual Scene Maps, which produces smoother but less semantically accurate transfers.
Nevertheless, our method attains a comparable length distortion score while maintaining small Waypoint AED and high inlier ratio values.

\begin{table*}[t]
\centering
\caption{
  Full results on 3D-FRONT and ARKitScenes including additional metrics.
  Paired split uses Single-GT annotations.
  SS denotes the single-side split.
  \colorbox{red!30}{\strut Red}, \colorbox{orange!40}{\strut orange}, and \colorbox{yellow!50}{\strut yellow} cells indicate the 1st, 2nd, and 3rd best result per column, respectively.
}
\label{tab:supp_full_results}
\resizebox{\textwidth}{!}{%
\setlength{\tabcolsep}{5pt}
\begin{tabular}{l ccc c| ccc c}
\toprule
& \multicolumn{3}{c}{3D-FRONT Paired} & \multicolumn{1}{c|}{3D-FRONT SS}
& \multicolumn{3}{c}{ARKit Paired} & \multicolumn{1}{c}{ARKit SS} \\
\cmidrule(lr){2-4}\cmidrule(lr){5-5}\cmidrule(lr){6-8}\cmidrule(lr){9-9}
Method
  & \makecell{Wpt.\\AED\,$\downarrow$}
  & \makecell{Inlier\\@1.0\,$\uparrow$}
  & \makecell{Inlier\\@2.0\,$\uparrow$}
  & \makecell{Len.\\Dist.\,$\downarrow$}
  & \makecell{Wpt.\\AED\,$\downarrow$}
  & \makecell{Inlier\\@1.0\,$\uparrow$}
  & \makecell{Inlier\\@2.0\,$\uparrow$}
  & \makecell{Len.\\Dist.\,$\downarrow$} \\
\midrule
Feature NN                            & 1.948 & 0.300 & 0.641 & 0.537 & 1.721 & 0.297 & 0.726 & 0.629 \\
Neural Contextual Scene Maps~\cite{}  & 1.877 & \rankthree{0.385} & 0.596 & \rankone{0.127} & 1.850 & 0.243 & 0.623 & \rankone{0.164} \\
Foundation Model Analogies            & 2.191 & 0.226 & 0.567 & 0.373 & 1.945 & 0.345 & 0.683 & 0.378 \\
Monte Carlo Localization              & 2.086 & 0.229 & 0.549 & 0.651 & 3.259 & 0.277 & 0.620 & 0.720 \\
Object Node Graph Match~\cite{}       & 1.697 & 0.310 & 0.667 & \rankthree{0.194} & 2.107 & 0.280 & 0.675 & \ranktwo{0.179} \\
Object Node Graph Match + ICP~\cite{} & \rankthree{1.595} & 0.321 & \rankthree{0.713} & 0.228 & 1.821 & 0.323 & \rankthree{0.767} & 0.274 \\
Keypoints Graph Match                 & 2.211 & 0.265 & 0.571 & \ranktwo{0.179} & 1.794 & \rankthree{0.364} & 0.744 & 0.338 \\
Multi-modal LLM (Dense)~\cite{}       & 2.019 & 0.259 & 0.626 & 0.316 & 1.817 & 0.308 & 0.713 & \rankthree{0.265} \\
Multi-modal LLM (Waypoint)~\cite{}    & 2.051 & 0.266 & 0.587 & 0.407 & 2.051 & 0.259 & 0.587 & 0.278 \\
\midrule
Ours (Dense)                          & \rankone{1.394} & \rankone{0.545} & \rankone{0.777} & 0.270 & \ranktwo{1.533} & \ranktwo{0.499} & \rankone{0.794} & 0.327 \\
Ours (Waypoint)                       & \ranktwo{1.492} & \ranktwo{0.489} & \ranktwo{0.741} & 0.272 & \rankone{1.347} & \rankone{0.523} & \ranktwo{0.793} & 0.329 \\
\bottomrule
\end{tabular}%
}
\end{table*}

\subsection{Additional Qualitative Results}

\begin{figure*}[t]
  \centering
  \includegraphics[width=\linewidth]{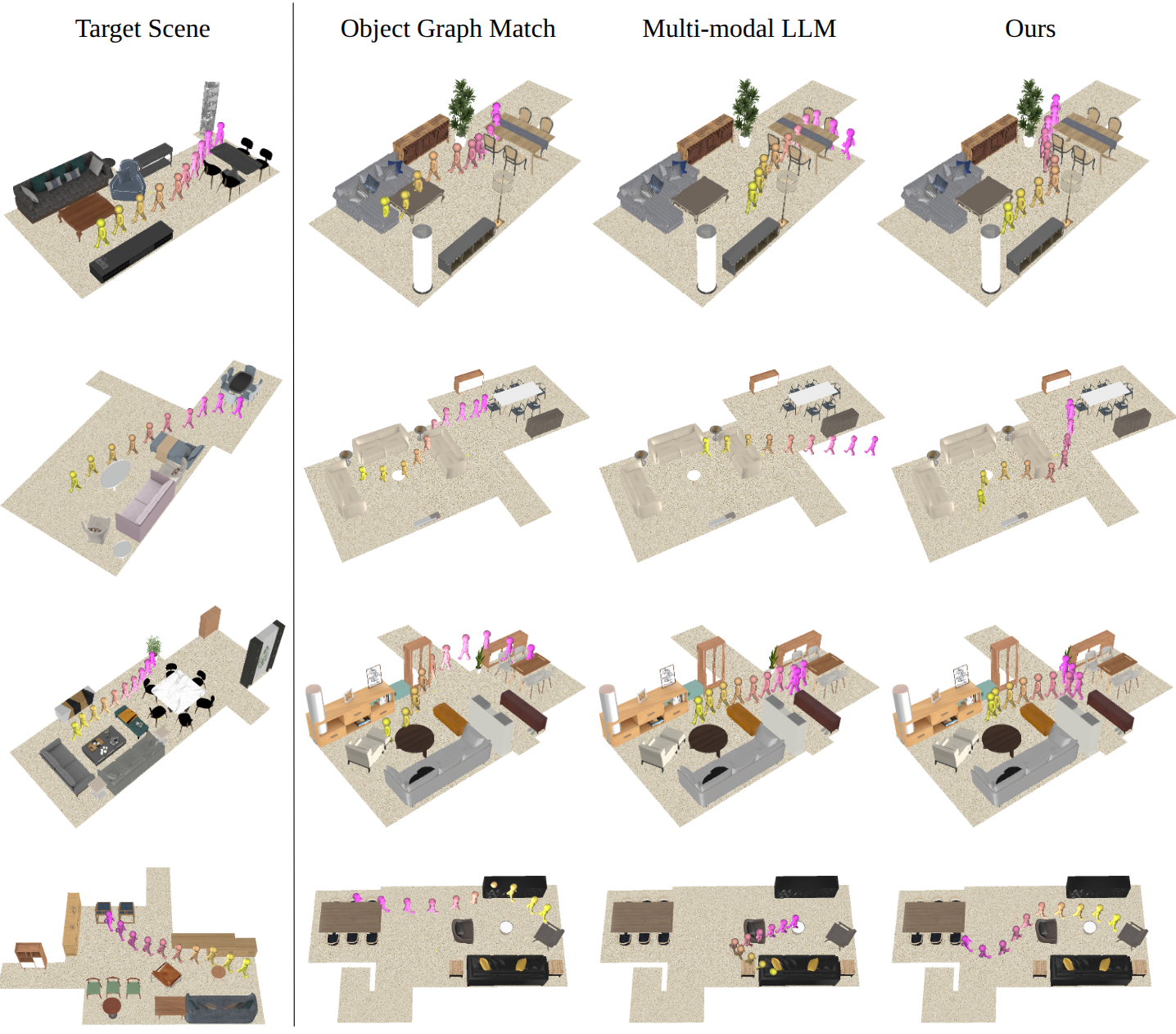}
  \caption{
    Additional qualitative results on 3D-FRONT scenes.
    Each row shows the target scene trajectory (leftmost) alongside results from Object Node Graph Match, Multi-modal LLM, and our method.
  }
  \label{fig:qualitative_tdfront_supp}
\end{figure*}

\begin{figure*}[t]
  \centering
  \includegraphics[width=\linewidth]{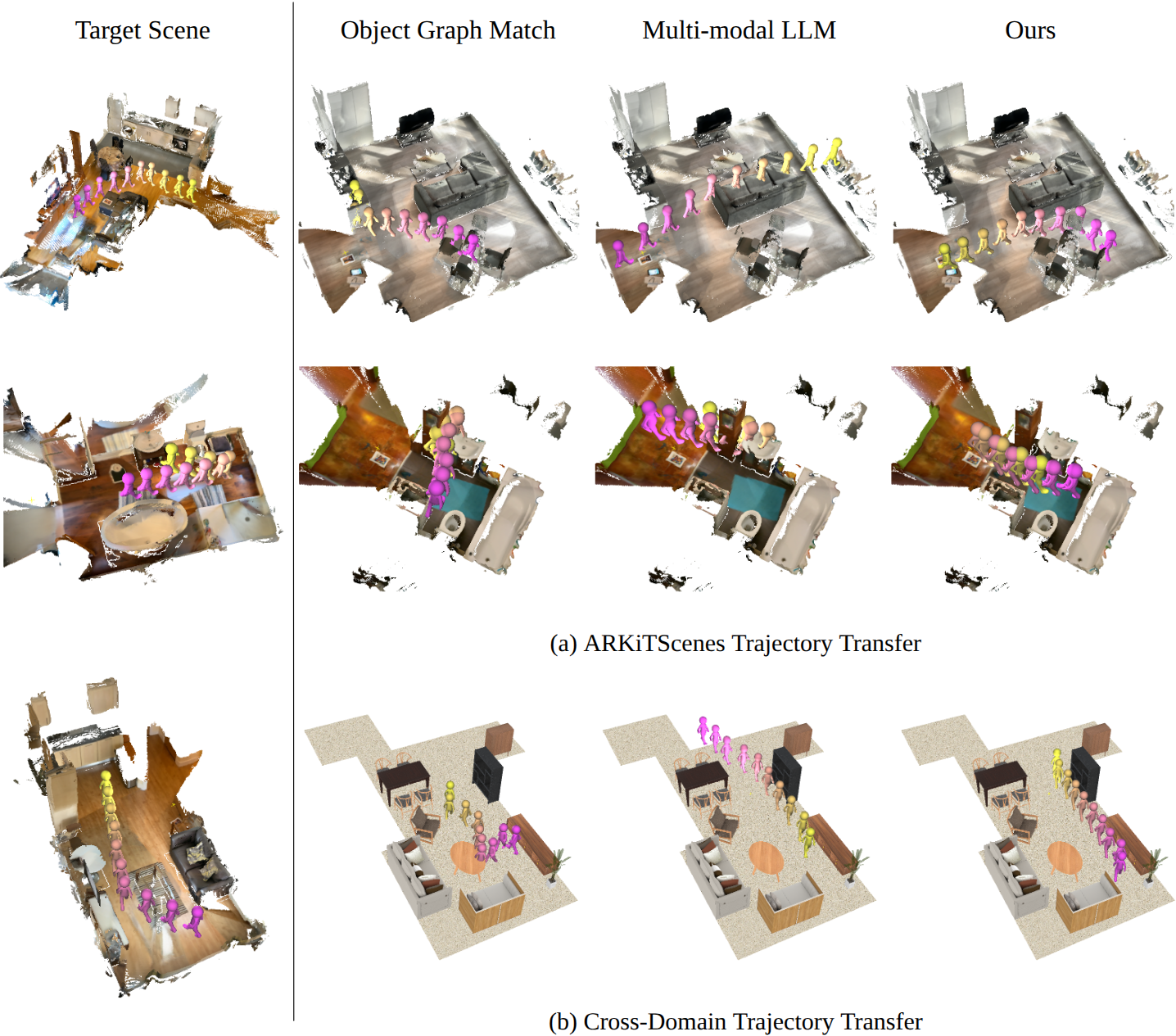}
  \caption{
    Additional qualitative results on ARKitScenes (top two rows) and cross-domain pairs (bottom row).
    Each row shows the target scene trajectory (leftmost) alongside results from Object Node Graph Match, Multi-modal LLM, and our method.
  }
  \label{fig:qualitative_arkit_supp}
\end{figure*}

Fig.~\ref{fig:qualitative_tdfront_supp} shows additional qualitative comparisons on 3D-FRONT scenes.
Fig.~\ref{fig:qualitative_arkit_supp} shows additional results on ARKitScenes and cross-domain pairs.
In both figures, our method produces trajectories that more faithfully follow the semantic layout of the reference scene compared to Object Node Graph Match and Multi-modal LLM baselines.
We also attach a video showing the applications of analogical trajectory transfer, \texttt{applications.mp4}.

\section{Application Implementation Details}
\label{sec:supp_app_impl}

\noindent\textbf{Multi-trajectory Transfer.}
A single smooth map is estimated for the scene pair and applied to produce an initial transfer of each agent trajectory.
Trajectory refinement is then applied separately for each agent, so that each trajectory is independently optimized to conform to the reference scene geometry.

\noindent\textbf{Virtual Co-presence.}
For each target scene, a smooth map is estimated to the shared reference scene.
The smooth maps are then used to produce an initial transfer for each of the target scene trajectories, followed by per-trajectory refinement.
The result is a set of trajectories, originating from different target scenes, that are all transferred to a common reference scene.

\noindent\textbf{Sim-to-Real Robot Motion Transfer.}
A smooth map from the synthetic source scene to the real-world target scene is estimated and used to produce an initial transfer of the human motion trajectory, followed by refinement.
The transferred human trajectory is retargeted to a robot trajectory by correcting for the height difference between the human and the robot.
The robot heading at each waypoint is determined by computing the angle of the tangent vector of the trajectory at that point.

\noindent\textbf{Camera Trajectory Transfer.}
A smooth map is estimated between the source and target scenes and used to produce an initial transfer of the camera path, followed by refinement.
The camera heading at each point along the transferred path is determined in a similar manner as sim-to-real robot motion transfer.

\section{Method Implementation Details}
\label{sec:supp_method_details}

\noindent\textbf{Affinity Matrix Construction.}
The affinity matrix $\mathbf{K} \in \mathbb{R}^{(N_\text{tgt} N_\text{ref}) \times (N_\text{tgt} N_\text{ref})}$ encodes both node-level feature similarity and edge-level geometric compatibility.
The diagonal entry $K_{pq, pq}$ is the inner product of the normalized 3D foundation model feature vectors~\citep{sonata} at target node $p$ and reference node $q$, measuring their semantic similarity.
The off-diagonal entry $K_{pq, p'q'}$ encodes the geometric compatibility of simultaneously matching the pairs $(p, q)$ and $(p', q')$, computed as the inverse absolute difference of the corresponding edge lengths:
$K_{pq, p'q'} = 1 / (|e_{pp'} - e_{qq'}| + \epsilon)$,
where $e_{pp'}$ and $e_{qq'}$ are the edge lengths between nodes $p, p'$ in the target graph and $q, q'$ in the reference graph, respectively, and $\epsilon$ is a small constant for numerical stability.

\noindent\textbf{Feature Loss and Spatial Neighborhood.}
The feature loss $E_\text{feat}$ (Sec.~\ref{sec:traj_transfer}) attracts a sparse subset $\mathbf{T}_\text{S}$ of trajectory points toward semantically matching positions in the reference scene.
For each sampled trajectory point $t_i \in \mathbf{T}_\text{S}$, the feature-matched target position $\tilde{v}_i$ is found by searching within a spatial disk of radius $r$ around the warped position $u_i = \phi_\text{global}(t_i)$ in the reference scene's XZ plane.
Among all reference scene points within this disk, the one whose 3D foundation model features best match those of $t_i$ is selected as $\tilde{v}_i$.
The search radius is set to $r = 1.0$\,m, and $|\mathbf{T}_\text{S}|$ is set to $50$ evenly-spaced samples along the trajectory.

\section{Additional Ablation Experiments}
\label{sec:supp_additional_ablation}
\begin{table}[t]
\centering
\caption{
  Ablation of cost terms in the smooth map assembly objective (Eq.~\ref{eq:assign_cost}) on 3D-FRONT Paired (Multiple-GT).
  $\checkmark$/$\times$ indicate whether a term is included.
  The best result per column is \textbf{bolded}.
  The last row corresponds to our full model.
}
\label{tab:supp_ablation_assembly}
\setlength{\tabcolsep}{6pt}
\begin{tabular}{ccc ccc}
\toprule
$\mathcal{L}_\text{feat}$ & $\mathcal{L}_\text{nav}$ & $\mathcal{L}_\text{distort}$
  & \makecell{Inlier\\@1.0\,$\uparrow$}
  & \makecell{Inlier\\@1.5\,$\uparrow$}
  & \makecell{Inlier\\@2.0\,$\uparrow$} \\
\midrule
$\checkmark$ & $\times$        & $\times$        & 0.484 & 0.603 & 0.696 \\
$\checkmark$ & $\checkmark$ & $\times$        & 0.554 & 0.714 & 0.815 \\
\midrule
$\checkmark$ & $\checkmark$ & $\checkmark$ & \textbf{0.576} & \textbf{0.743} & \textbf{0.852} \\
\bottomrule
\end{tabular}
\end{table}

\begin{table}[t]
\centering
\caption{
  Ablation of cost terms in the trajectory refinement objective (Eq.~\ref{eq:traj_warp_cost}) on 3D-FRONT Paired (Multiple-GT).
  $\checkmark$/$\times$ indicate whether a term is included.
  The best result per column is \textbf{bolded}.
  The last row corresponds to our full model.
}
\label{tab:supp_ablation_traj}
\setlength{\tabcolsep}{6pt}
\begin{tabular}{cccc ccc}
\toprule
$E_\text{feat}$ & $E_\text{nav}$ & $E_\text{anchor}$ & $E_\text{shape}$
  & \makecell{Inlier\\@1.0\,$\uparrow$}
  & \makecell{Inlier\\@1.5\,$\uparrow$}
  & \makecell{Inlier\\@2.0\,$\uparrow$} \\
\midrule
$\checkmark$ & $\times$        & $\times$        & $\times$        & 0.572 & 0.736 & 0.806 \\
$\checkmark$ & $\checkmark$ & $\times$        & $\times$        & 0.554 & 0.737 & 0.809 \\
$\checkmark$ & $\checkmark$ & $\checkmark$ & $\times$        & 0.567 & 0.740 & 0.808 \\
\midrule
$\checkmark$ & $\checkmark$ & $\checkmark$ & $\checkmark$ & \textbf{0.576} & \textbf{0.743} & \textbf{0.852} \\
\bottomrule
\end{tabular}
\end{table}

\noindent\textbf{Sensitivity to Smooth Map Assembly Weights.}
We ablate the cost terms $\mathcal{L}_\text{feat}$, $\mathcal{L}_\text{nav}$, and $\mathcal{L}_\text{distort}$ in the smooth map assembly objective (Eq.~\ref{eq:assign_cost}) on 3D-FRONT Paired (Multiple-GT).
Tab.~\ref{tab:supp_ablation_assembly} shows that the three terms combined show the best performance.

\begin{table}[t]
\centering
\caption{
  Extended feature backbone comparison on 3D-FRONT Paired (Single-GT).
  The best result per column is \textbf{bolded}.
  The last row corresponds to our default model using Sonata~\citep{sonata} features.
}
\label{tab:supp_ablation_feat}
\setlength{\tabcolsep}{6pt}
\begin{tabular}{l cccc}
\toprule
Method
  & \makecell{Traj.\\AED\,$\downarrow$}
  & \makecell{Inlier\\@1.0\,$\uparrow$}
  & \makecell{Inlier\\@1.5\,$\uparrow$}
  & \makecell{Inlier\\@2.0\,$\uparrow$} \\
\midrule
Ours w/ DINOv2~\citep{dinov2}           & 1.835 & 0.248 & 0.387 & 0.551 \\
Ours w/ Vector Neurons~\citep{vector_neuron}         & 1.961 & 0.274 & 0.419 & 0.575 \\
Ours w/ Concerto~\citep{concerto}       & 1.143 & 0.532 & 0.657 & 0.762 \\
Ours w/ Utonia~\citep{utonia}           & 1.389 & 0.402 & 0.579 & 0.701 \\
\midrule
Ours (Dense)                            & \textbf{0.997} & \textbf{0.545} & \textbf{0.672} & \textbf{0.777} \\
\bottomrule
\end{tabular}
\end{table}

\noindent\textbf{Sensitivity to Trajectory Refinement Weights.}
We ablate the cost terms $E_\text{feat}$, $E_\text{nav}$, $E_\text{anchor}$, and $E_\text{shape}$ in the trajectory refinement objective (Eq.~\ref{eq:traj_warp_cost}) on 3D-FRONT Paired (Multiple-GT).
Tab.~\ref{tab:supp_ablation_traj} shows that the full combination of all four terms yields the best results.

\begin{table}[t]
\centering
\caption{
  Sensitivity to the target cluster size $N_\text{cluster}$ on 3D-FRONT Paired (Multiple-GT).
  The best result per column is \textbf{bolded}.
  The row with $N_\text{cluster} = 4$ corresponds to our default setting.
  The last row replaces agglomerative clustering with an MLLM that groups objects based on top-down renderings.
}
\label{tab:supp_ablation_cluster}
\setlength{\tabcolsep}{8pt}
\begin{tabular}{l ccc}
\toprule
Method
  & \makecell{Inlier\\@1.0\,$\uparrow$}
  & \makecell{Inlier\\@1.5\,$\uparrow$}
  & \makecell{Inlier\\@2.0\,$\uparrow$} \\
\midrule
$N_\text{cluster} = 3$ & 0.599 & 0.758 & 0.821 \\
$N_\text{cluster} = 4$ & 0.576 & 0.743 & 0.852 \\
$N_\text{cluster} = 5$ & \textbf{0.605} & \textbf{0.763} & 0.846 \\
$N_\text{cluster} = 6$ & 0.583 & 0.726 & \textbf{0.857} \\
\midrule
\small{MLLM-based Clustering} & 0.592 & 0.720 & 0.806 \\
\bottomrule
\end{tabular}
\end{table}

\noindent\textbf{Extended Feature Backbone Comparison.}
We extend the feature backbone ablation in Sec.~\ref{sec:ablation} to Tab.~\ref{tab:supp_ablation_feat}, which includes an additional 2D foundation model feature (DINOv2~\citep{dinov2}), rotation-equivariant point cloud feature descriptors (Vector Neurons~\citep{vector_neuron}), and other 3D foundation model features for point cloud understanding (Concerto~\citep{concerto}, and Utonia~\citep{utonia}).
Our default model using Sonata~\citep{sonata} features shows the best performance, which supports our design choice of using Sonata as the default feature descriptor for our method.
Interestingly, while Concerto~\citep{concerto} and Utonia~\citep{utonia} are follow-up works of Sonata~\citep{sonata}, we find that Sonata features perform best in our task.
One possible explanation is that while training on diverse point cloud data as in Utonia~\citep{utonia} or learning a joint 2D-3D embedding space as in Concerto~\citep{concerto} may help in tasks such as instance or semantic segmentation, the performance gains for analogical reasoning are less affected by these modifications.
A detailed investigation on the relationship between 3D foundation model features and the capacity to perform analogical reasoning warrants future work.

\noindent\textbf{Sensitivity to Clustering Parameters.}
We ablate the target cluster size $N_\text{cluster}$, which controls how many object instances are grouped into each cluster, with a default of $N_\text{cluster} = 4$.
Tab.~\ref{tab:supp_ablation_cluster} reports performance for $N_\text{cluster} \in \{3, 4, 5, 6\}$, and shows that performance is relatively stable across this range.
We additionally compare against a multi-modal LLM (MLLM)-based clustering alternative, in which a multi-modal LLM~\citep{gpt5} is shown a top-down rendering of each scene and prompted to group objects into clusters based on their functionality and semantics.
Our method shows stable performance regardless of the clustering method used, suggesting that the overall pipeline is robust to the choice of object grouping strategy.


\newpage
\section*{NeurIPS Paper Checklist}

The checklist is designed to encourage best practices for responsible machine learning research, addressing issues of reproducibility, transparency, research ethics, and societal impact. Do not remove the checklist: {\bf The papers not including the checklist will be desk rejected.} The checklist should follow the references and follow the (optional) supplemental material.  The checklist does NOT count towards the page
limit. 

Please read the checklist guidelines carefully for information on how to answer these questions. For each question in the checklist:
\begin{itemize}
    \item You should answer \answerYes{}, \answerNo{}, or \answerNA{}.
    \item \answerNA{} means either that the question is Not Applicable for that particular paper or the relevant information is Not Available.
    \item Please provide a short (1--2 sentence) justification right after your answer (even for \answerNA). 
\end{itemize}

{\bf The checklist answers are an integral part of your paper submission.} They are visible to the reviewers, area chairs, senior area chairs, and ethics reviewers. You will also be asked to include it (after eventual revisions) with the final version of your paper, and its final version will be published with the paper.

The reviewers of your paper will be asked to use the checklist as one of the factors in their evaluation. While \answerYes{} is generally preferable to \answerNo{}, it is perfectly acceptable to answer \answerNo{} provided a proper justification is given (e.g., error bars are not reported because it would be too computationally expensive'' or ``we were unable to find the license for the dataset we used''). In general, answering \answerNo{} or \answerNA{} is not grounds for rejection. While the questions are phrased in a binary way, we acknowledge that the true answer is often more nuanced, so please just use your best judgment and write a justification to elaborate. All supporting evidence can appear either in the main paper or the supplemental material, provided in appendix. If you answer \answerYes{} to a question, in the justification please point to the section(s) where related material for the question can be found.

IMPORTANT, please:
\begin{itemize}
    \item {\bf Delete this instruction block, but keep the section heading ``NeurIPS Paper Checklist"},
    \item  {\bf Keep the checklist subsection headings, questions/answers and guidelines below.}
    \item {\bf Do not modify the questions and only use the provided macros for your answers}.
\end{itemize}


\begin{enumerate}

\item {\bf Claims}
    \item[] Question: Do the main claims made in the abstract and introduction accurately reflect the paper's contributions and scope?
    \item[] Answer: \answerYes{} 
    \item[] Justification: Our paper's contributions are in solving a new task called analogical trajectory transfer, and the abstract and introduction reflects this.
    \item[] Guidelines:
    \begin{itemize}
        \item The answer \answerNA{} means that the abstract and introduction do not include the claims made in the paper.
        \item The abstract and/or introduction should clearly state the claims made, including the contributions made in the paper and important assumptions and limitations. A \answerNo{} or \answerNA{} answer to this question will not be perceived well by the reviewers. 
        \item The claims made should match theoretical and experimental results, and reflect how much the results can be expected to generalize to other settings. 
        \item It is fine to include aspirational goals as motivation as long as it is clear that these goals are not attained by the paper. 
    \end{itemize}

\item {\bf Limitations}
    \item[] Question: Does the paper discuss the limitations of the work performed by the authors?
    \item[] Answer: \answerYes{} 
    \item[] Justification: The limitations are discussed in the Conclusion section.
    \item[] Guidelines:
    \begin{itemize}
        \item The answer \answerNA{} means that the paper has no limitation while the answer \answerNo{} means that the paper has limitations, but those are not discussed in the paper. 
        \item The authors are encouraged to create a separate ``Limitations'' section in their paper.
        \item The paper should point out any strong assumptions and how robust the results are to violations of these assumptions (e.g., independence assumptions, noiseless settings, model well-specification, asymptotic approximations only holding locally). The authors should reflect on how these assumptions might be violated in practice and what the implications would be.
        \item The authors should reflect on the scope of the claims made, e.g., if the approach was only tested on a few datasets or with a few runs. In general, empirical results often depend on implicit assumptions, which should be articulated.
        \item The authors should reflect on the factors that influence the performance of the approach. For example, a facial recognition algorithm may perform poorly when image resolution is low or images are taken in low lighting. Or a speech-to-text system might not be used reliably to provide closed captions for online lectures because it fails to handle technical jargon.
        \item The authors should discuss the computational efficiency of the proposed algorithms and how they scale with dataset size.
        \item If applicable, the authors should discuss possible limitations of their approach to address problems of privacy and fairness.
        \item While the authors might fear that complete honesty about limitations might be used by reviewers as grounds for rejection, a worse outcome might be that reviewers discover limitations that aren't acknowledged in the paper. The authors should use their best judgment and recognize that individual actions in favor of transparency play an important role in developing norms that preserve the integrity of the community. Reviewers will be specifically instructed to not penalize honesty concerning limitations.
    \end{itemize}

\item {\bf Theory assumptions and proofs}
    \item[] Question: For each theoretical result, does the paper provide the full set of assumptions and a complete (and correct) proof?
    \item[] Answer: \answerNA{} 
    \item[] Justification: Our paper does not have any theorectical proofs.
    \item[] Guidelines:
    \begin{itemize}
        \item The answer \answerNA{} means that the paper does not include theoretical results. 
        \item All the theorems, formulas, and proofs in the paper should be numbered and cross-referenced.
        \item All assumptions should be clearly stated or referenced in the statement of any theorems.
        \item The proofs can either appear in the main paper or the supplemental material, but if they appear in the supplemental material, the authors are encouraged to provide a short proof sketch to provide intuition. 
        \item Inversely, any informal proof provided in the core of the paper should be complemented by formal proofs provided in appendix or supplemental material.
        \item Theorems and Lemmas that the proof relies upon should be properly referenced. 
    \end{itemize}

    \item {\bf Experimental result reproducibility}
    \item[] Question: Does the paper fully disclose all the information needed to reproduce the main experimental results of the paper to the extent that it affects the main claims and/or conclusions of the paper (regardless of whether the code and data are provided or not)?
    \item[] Answer: \answerYes{} 
    \item[] Justification: We provide detailed implementation information in the supplementary material.
    \item[] Guidelines:
    \begin{itemize}
        \item The answer \answerNA{} means that the paper does not include experiments.
        \item If the paper includes experiments, a \answerNo{} answer to this question will not be perceived well by the reviewers: Making the paper reproducible is important, regardless of whether the code and data are provided or not.
        \item If the contribution is a dataset and\slash or model, the authors should describe the steps taken to make their results reproducible or verifiable. 
        \item Depending on the contribution, reproducibility can be accomplished in various ways. For example, if the contribution is a novel architecture, describing the architecture fully might suffice, or if the contribution is a specific model and empirical evaluation, it may be necessary to either make it possible for others to replicate the model with the same dataset, or provide access to the model. In general. releasing code and data is often one good way to accomplish this, but reproducibility can also be provided via detailed instructions for how to replicate the results, access to a hosted model (e.g., in the case of a large language model), releasing of a model checkpoint, or other means that are appropriate to the research performed.
        \item While NeurIPS does not require releasing code, the conference does require all submissions to provide some reasonable avenue for reproducibility, which may depend on the nature of the contribution. For example
        \begin{enumerate}
            \item If the contribution is primarily a new algorithm, the paper should make it clear how to reproduce that algorithm.
            \item If the contribution is primarily a new model architecture, the paper should describe the architecture clearly and fully.
            \item If the contribution is a new model (e.g., a large language model), then there should either be a way to access this model for reproducing the results or a way to reproduce the model (e.g., with an open-source dataset or instructions for how to construct the dataset).
            \item We recognize that reproducibility may be tricky in some cases, in which case authors are welcome to describe the particular way they provide for reproducibility. In the case of closed-source models, it may be that access to the model is limited in some way (e.g., to registered users), but it should be possible for other researchers to have some path to reproducing or verifying the results.
        \end{enumerate}
    \end{itemize}

\item {\bf Open access to data and code}
    \item[] Question: Does the paper provide open access to the data and code, with sufficient instructions to faithfully reproduce the main experimental results, as described in supplemental material?
    \item[] Answer: \answerNo{} 
    \item[] Justification: While we do not provide access to data and code at this point, we will release the data and code upon acceptance.
    \item[] Guidelines:
    \begin{itemize}
        \item The answer \answerNA{} means that paper does not include experiments requiring code.
        \item Please see the NeurIPS code and data submission guidelines (\url{https://neurips.cc/public/guides/CodeSubmissionPolicy}) for more details.
        \item While we encourage the release of code and data, we understand that this might not be possible, so \answerNo{} is an acceptable answer. Papers cannot be rejected simply for not including code, unless this is central to the contribution (e.g., for a new open-source benchmark).
        \item The instructions should contain the exact command and environment needed to run to reproduce the results. See the NeurIPS code and data submission guidelines (\url{https://neurips.cc/public/guides/CodeSubmissionPolicy}) for more details.
        \item The authors should provide instructions on data access and preparation, including how to access the raw data, preprocessed data, intermediate data, and generated data, etc.
        \item The authors should provide scripts to reproduce all experimental results for the new proposed method and baselines. If only a subset of experiments are reproducible, they should state which ones are omitted from the script and why.
        \item At submission time, to preserve anonymity, the authors should release anonymized versions (if applicable).
        \item Providing as much information as possible in supplemental material (appended to the paper) is recommended, but including URLs to data and code is permitted.
    \end{itemize}

\item {\bf Experimental setting/details}
    \item[] Question: Does the paper specify all the training and test details (e.g., data splits, hyperparameters, how they were chosen, type of optimizer) necessary to understand the results?
    \item[] Answer: \answerYes{} 
    \item[] Justification: The experiment settings are specified in the supplementary material.
    \item[] Guidelines:
    \begin{itemize}
        \item The answer \answerNA{} means that the paper does not include experiments.
        \item The experimental setting should be presented in the core of the paper to a level of detail that is necessary to appreciate the results and make sense of them.
        \item The full details can be provided either with the code, in appendix, or as supplemental material.
    \end{itemize}

\item {\bf Experiment statistical significance}
    \item[] Question: Does the paper report error bars suitably and correctly defined or other appropriate information about the statistical significance of the experiments?
    \item[] Answer: \answerNo{} 
    \item[] Justification: Our paper does not include error bars.
    \item[] Guidelines:
    \begin{itemize}
        \item The answer \answerNA{} means that the paper does not include experiments.
        \item The authors should answer \answerYes{} if the results are accompanied by error bars, confidence intervals, or statistical significance tests, at least for the experiments that support the main claims of the paper.
        \item The factors of variability that the error bars are capturing should be clearly stated (for example, train/test split, initialization, random drawing of some parameter, or overall run with given experimental conditions).
        \item The method for calculating the error bars should be explained (closed form formula, call to a library function, bootstrap, etc.)
        \item The assumptions made should be given (e.g., Normally distributed errors).
        \item It should be clear whether the error bar is the standard deviation or the standard error of the mean.
        \item It is OK to report 1-sigma error bars, but one should state it. The authors should preferably report a 2-sigma error bar than state that they have a 96\% CI, if the hypothesis of Normality of errors is not verified.
        \item For asymmetric distributions, the authors should be careful not to show in tables or figures symmetric error bars that would yield results that are out of range (e.g., negative error rates).
        \item If error bars are reported in tables or plots, the authors should explain in the text how they were calculated and reference the corresponding figures or tables in the text.
    \end{itemize}

\item {\bf Experiments compute resources}
    \item[] Question: For each experiment, does the paper provide sufficient information on the computer resources (type of compute workers, memory, time of execution) needed to reproduce the experiments?
    \item[] Answer: \answerYes{} 
    \item[] Justification: Our method does not require training, and thus does not require a large number of GPUs. All of our experiments are conduced using RTX 3090 and A6000 GPUs.
    \item[] Guidelines:
    \begin{itemize}
        \item The answer \answerNA{} means that the paper does not include experiments.
        \item The paper should indicate the type of compute workers CPU or GPU, internal cluster, or cloud provider, including relevant memory and storage.
        \item The paper should provide the amount of compute required for each of the individual experimental runs as well as estimate the total compute. 
        \item The paper should disclose whether the full research project required more compute than the experiments reported in the paper (e.g., preliminary or failed experiments that didn't make it into the paper). 
    \end{itemize}
    
\item {\bf Code of ethics}
    \item[] Question: Does the research conducted in the paper conform, in every respect, with the NeurIPS Code of Ethics \url{https://neurips.cc/public/EthicsGuidelines}?
    \item[] Answer: \answerYes{} 
    \item[] Justification: We have checked the NeurIPS code of ethics, and verify that our paper complies with the code of ethics.
    \item[] Guidelines:
    \begin{itemize}
        \item The answer \answerNA{} means that the authors have not reviewed the NeurIPS Code of Ethics.
        \item If the authors answer \answerNo, they should explain the special circumstances that require a deviation from the Code of Ethics.
        \item The authors should make sure to preserve anonymity (e.g., if there is a special consideration due to laws or regulations in their jurisdiction).
    \end{itemize}

\item {\bf Broader impacts}
    \item[] Question: Does the paper discuss both potential positive societal impacts and negative societal impacts of the work performed?
    \item[] Answer: \answerYes{} 
    \item[] Justification: Our paper includes discussions on possible application areas of our newly proposed task, analogical trajectory transfer.
    \item[] Guidelines:
    \begin{itemize}
        \item The answer \answerNA{} means that there is no societal impact of the work performed.
        \item If the authors answer \answerNA{} or \answerNo, they should explain why their work has no societal impact or why the paper does not address societal impact.
        \item Examples of negative societal impacts include potential malicious or unintended uses (e.g., disinformation, generating fake profiles, surveillance), fairness considerations (e.g., deployment of technologies that could make decisions that unfairly impact specific groups), privacy considerations, and security considerations.
        \item The conference expects that many papers will be foundational research and not tied to particular applications, let alone deployments. However, if there is a direct path to any negative applications, the authors should point it out. For example, it is legitimate to point out that an improvement in the quality of generative models could be used to generate Deepfakes for disinformation. On the other hand, it is not needed to point out that a generic algorithm for optimizing neural networks could enable people to train models that generate Deepfakes faster.
        \item The authors should consider possible harms that could arise when the technology is being used as intended and functioning correctly, harms that could arise when the technology is being used as intended but gives incorrect results, and harms following from (intentional or unintentional) misuse of the technology.
        \item If there are negative societal impacts, the authors could also discuss possible mitigation strategies (e.g., gated release of models, providing defenses in addition to attacks, mechanisms for monitoring misuse, mechanisms to monitor how a system learns from feedback over time, improving the efficiency and accessibility of ML).
    \end{itemize}
    
\item {\bf Safeguards}
    \item[] Question: Does the paper describe safeguards that have been put in place for responsible release of data or models that have a high risk for misuse (e.g., pre-trained language models, image generators, or scraped datasets)?
    \item[] Answer: \answerNA{} 
    \item[] Justification: Our work does not use data or models that have high risk for misuse.
    \item[] Guidelines:
    \begin{itemize}
        \item The answer \answerNA{} means that the paper poses no such risks.
        \item Released models that have a high risk for misuse or dual-use should be released with necessary safeguards to allow for controlled use of the model, for example by requiring that users adhere to usage guidelines or restrictions to access the model or implementing safety filters. 
        \item Datasets that have been scraped from the Internet could pose safety risks. The authors should describe how they avoided releasing unsafe images.
        \item We recognize that providing effective safeguards is challenging, and many papers do not require this, but we encourage authors to take this into account and make a best faith effort.
    \end{itemize}

\item {\bf Licenses for existing assets}
    \item[] Question: Are the creators or original owners of assets (e.g., code, data, models), used in the paper, properly credited and are the license and terms of use explicitly mentioned and properly respected?
    \item[] Answer: \answerYes{} 
    \item[] Justification: The creators of prior works that we use as baselines and datasets are properly credited in our paper.
    \item[] Guidelines:
    \begin{itemize}
        \item The answer \answerNA{} means that the paper does not use existing assets.
        \item The authors should cite the original paper that produced the code package or dataset.
        \item The authors should state which version of the asset is used and, if possible, include a URL.
        \item The name of the license (e.g., CC-BY 4.0) should be included for each asset.
        \item For scraped data from a particular source (e.g., website), the copyright and terms of service of that source should be provided.
        \item If assets are released, the license, copyright information, and terms of use in the package should be provided. For popular datasets, \url{paperswithcode.com/datasets} has curated licenses for some datasets. Their licensing guide can help determine the license of a dataset.
        \item For existing datasets that are re-packaged, both the original license and the license of the derived asset (if it has changed) should be provided.
        \item If this information is not available online, the authors are encouraged to reach out to the asset's creators.
    \end{itemize}

\item {\bf New assets}
    \item[] Question: Are new assets introduced in the paper well documented and is the documentation provided alongside the assets?
    \item[] Answer: \answerNA{} 
    \item[] Justification: Our paper does not release any new assets at this point.
    \item[] Guidelines:
    \begin{itemize}
        \item The answer \answerNA{} means that the paper does not release new assets.
        \item Researchers should communicate the details of the dataset\slash code\slash model as part of their submissions via structured templates. This includes details about training, license, limitations, etc. 
        \item The paper should discuss whether and how consent was obtained from people whose asset is used.
        \item At submission time, remember to anonymize your assets (if applicable). You can either create an anonymized URL or include an anonymized zip file.
    \end{itemize}

\item {\bf Crowdsourcing and research with human subjects}
    \item[] Question: For crowdsourcing experiments and research with human subjects, does the paper include the full text of instructions given to participants and screenshots, if applicable, as well as details about compensation (if any)? 
    \item[] Answer: \answerNA{} 
    \item[] Justification: Our paper does not involve crowdsourcing nor research with human subjects.
    \item[] Guidelines:
    \begin{itemize}
        \item The answer \answerNA{} means that the paper does not involve crowdsourcing nor research with human subjects.
        \item Including this information in the supplemental material is fine, but if the main contribution of the paper involves human subjects, then as much detail as possible should be included in the main paper. 
        \item According to the NeurIPS Code of Ethics, workers involved in data collection, curation, or other labor should be paid at least the minimum wage in the country of the data collector. 
    \end{itemize}

\item {\bf Institutional review board (IRB) approvals or equivalent for research with human subjects}
    \item[] Question: Does the paper describe potential risks incurred by study participants, whether such risks were disclosed to the subjects, and whether Institutional Review Board (IRB) approvals (or an equivalent approval/review based on the requirements of your country or institution) were obtained?
    \item[] Answer: \answerNA{} 
    \item[] Justification: Our paper does not involve crowdsourcing nor research with human subjects.
    \item[] Guidelines:
    \begin{itemize}
        \item The answer \answerNA{} means that the paper does not involve crowdsourcing nor research with human subjects.
        \item Depending on the country in which research is conducted, IRB approval (or equivalent) may be required for any human subjects research. If you obtained IRB approval, you should clearly state this in the paper. 
        \item We recognize that the procedures for this may vary significantly between institutions and locations, and we expect authors to adhere to the NeurIPS Code of Ethics and the guidelines for their institution. 
        \item For initial submissions, do not include any information that would break anonymity (if applicable), such as the institution conducting the review.
    \end{itemize}

\item {\bf Declaration of LLM usage}
    \item[] Question: Does the paper describe the usage of LLMs if it is an important, original, or non-standard component of the core methods in this research? Note that if the LLM is used only for writing, editing, or formatting purposes and does \emph{not} impact the core methodology, scientific rigor, or originality of the research, declaration is not required.
    \item[] Answer: \answerNo{} 
    \item[] Justification: LLMs are only used for writing and editing assistance.
    \item[] Guidelines:
    \begin{itemize}
        \item The answer \answerNA{} means that the core method development in this research does not involve LLMs as any important, original, or non-standard components.
        \item Please refer to our LLM policy in the NeurIPS handbook for what should or should not be described.
    \end{itemize}

\end{enumerate}

\end{document}